\documentclass{article}

\usepackage{arxiv}
\usepackage{amsmath,latexsym,amssymb,amsfonts}
\usepackage[utf8]{inputenc} 
\usepackage[T1]{fontenc}    
\usepackage{hyperref}       
\usepackage{algorithmic}
\usepackage{tikz}
\usepackage[affil-it]{authblk}
\usepackage{amsthm}
\usepackage{caption}
\usetikzlibrary{fadings, positioning, arrows, shapes, shapes.misc, fit, calc}
\usepackage{subfigure}
\usepackage{graphicx}
\usepackage{float}
\usepackage{bm}
\usepackage{csquotes}
\usepackage{listings}
\usepackage{balance}
\usepackage{array}
\usepackage{amssymb}
\usepackage{hyperref}
\usepackage{mathtools}
\usepackage{url}

\usepackage{makecell}
\usepackage[algo2e,ruled,vlined,linesnumbered]{algorithm2e}
\usepackage[nodisplayskipstretch]{setspace}

\newtheorem{prop}{Proposition}
\newtheorem{remark}{Remark}

\usepackage{pgfplots}
\usepackage{siunitx}  
\newcommand{\RNum}[1]{\uppercase\expandafter{\romannumeral #1\relax}}

\title{Multigoal-oriented dual-weighted-residual error estimation using deep neural networks}

\author[4]{Ayan Chakraborty}
\author[5,6,*]{Thomas Wick}
\author[3,4,**]{Xiaoying Zhuang}
\author[1,2]{Timon Rabczuk}
\affil[1]{Division of Computational Mechanics, Ton Duc Thang University, Ho Chi Minh City, 99423, Vietnam}
\affil[2]{Faculty of Civil Engineering, Ton Duc Thang University, Ho Chi Minh City, 99423, Vietnam}
\affil[3]{Department of Geotechnical Engineering,College of Civil Engineering, Tongji University, Shanghai, China}
\affil[4]{Chair of Computational Science and Simulation Technology, Institute of Photonics, Faculty of Mathematics and
Physics,Leibniz University, Hannover, 30163, Germany}
\affil[5]{Institute of Applied Mathematics Welfengarten 1, AG Wissenschaftliches Rechnen,Leibniz
University,, Hannover, 30167, Germany}
\affil[6]{Cluster of Excellence PhoenixD (Photonics, Optics, and Engineering - Innovation Across Disciplines), Leibniz University
Hannover, Germany}
\affil[*]{Corresponding author: thomas.wick@ifam.uni-hannover.de}
\affil[**]{Corresponding author: zhuang@iop.uni-hannover.de}
\begin{document}

\maketitle

\begin{abstract}
Deep learning has shown successful application in visual recognition and certain artificial intelligence tasks. Deep learning is also considered as a powerful tool with high flexibility to approximate functions. In the present work, functions with desired properties are devised to approximate the solutions of PDEs. Our approach is based on a posteriori error estimation in which the adjoint problem is solved for the error localization to formulate an error estimator within the framework of neural network. An efficient and easy to implement algorithm is developed to obtain a posteriori error estimate for multiple goal functionals by employing the dual-weighted residual approach, which is followed by the computation of both primal and adjoint solutions using the neural network. The present study shows that such a data-driven model based learning has superior approximation of quantities of interest even with relatively less training data. The novel algorithmic developments are substantiated with numerical test examples. The advantages of using deep neural network over the shallow neural network are demonstrated and the convergence enhancing techniques are also presented.
\end{abstract}


\section{Introduction}
Differential equations are the cornerstone of a vast range of engineering fields and applied sciences. It is the most common class of simulation problem in science and engineering and is devised as a mathematical tool to study the physical world in a precise manner. So far a wide varieties of 
methods have been developed for solving differential equations, for instance  see~\cite{hughes2012finite,rabczuk2019extended,chakraborty2019finite,jia2019adaptive} and the references therein. In most of the previous works, solving PDEs is restricted to solve a linear system of algebraic equations which result from the discretization of the computational domain. In particular approximating the solution is a crucial step and most of such contributions intend to explore variety of such possibilities.\\
In this paper, we view the problem from a different angle. We propose a promising method for solving PDEs that entirely relies on the function approximation capabilities of the feed forward neural networks that has achieved immense success in approximation theory and set the foundations for a new paradigm in modelling as well as computations which also enriches deep learning with the longstanding developments and have made a transformative impact on modern technology. The recent breakthroughs in machine learning in the form of Deep Neural Network (DNN) have led to outstanding achievements in several areas and hence become extremely popular in computational sciences in recent years. Starting from visual recognition, image processing, natural language processing and in many other diversity,  DNNs with the capability of holding  large number of hidden layers has proven an extremely successful devise to approximate the target function. Thanks to the development of very powerful computational tools viz. Tensorflow~\cite{abadi2016tensorflow}, Pytorch~\cite{paszke2017automatic} libraries that provide the building blocks to train the machine with respect to different problems and are currently ones main focus in the field of applied mathematics and computer science.\\
Recent trends of applying DNN to approximate solutions are being increasingly used in the context of numerical computations of PDEs~\cite{samaniego2020energy,sirignano2018dgm,raissi2017physics,mishra2018machine}. With the explosive growth of computational resources, deep learning becomes rapidly evolving field. In the following, we will provide a brief overview. In~\cite{brunton2016discovering,rudy2017data,schaeffer2017learning}   sequential threshold Ridge regression is proposed, i.e. Lasso regression to identify PDE from data. In~\cite{xu2019dl}, so called  DL-PDE is proposed that combines neural network and  data driven discovery via sparse regression. Long et al.~\cite{long2018pde} proposed PDE-Net to learn differential operators  by employing a convolutional neural network. Berg and Nystrom~\cite{berg2019data} proposed an algorithm for discovering PDE from complex data set, however their approach requires a large amount of data and is robust in some specific cases. Beck et al.~\cite{beck2019machine} developed an algorithm for solving a class of quasilinear PDEs with high accuracy and speed. In addition, the authors in ~\cite{chaudhari2018deep} have proposed a scheme to optimize deep neural networks using PDEs.\\

In this paper, we consider the strong form of PDEs, leading to a collocation based approach. Given the fact that the learning process of DNN can be considered as an iterative process of minimizing the loss function, therefore it seems natural to consider the residual of the system as an ideal candidate for this function. Our main aim is to develop goal-oriented a posteriori error estimation with two objectives in mind. At one hand, we focus on the dual-weighted residual method \cite{BeRa96,becker2001optimal,AinsworthOden:2000,bangerth2003adaptive,braack_ern_2003,RiWi15_dwr,Od18}
and on the other hand, we consider multiple functionals of interest \cite{hartmann2003goal,hartmann2008multitarget,BruZhuZwie16,endtmayer2017partition,endtmayer2019multigoal,KerPruChaLaf2017,Endt21,Wi20_book}. These error estimators are formulated for nonlinear problems in which both the PDE and the functionals are nonlinear.  

The main objective of this work is to study such multigoal-oriented 
error estimators in the framework of machine learning. We mainly focus on the Deep Collocation Method (DCM) and we have shown that our method is capable of achieving a good prediction accuracy given a sufficient number of collocation points on the domain with sufficiently expressive neural network architecture. Our network is fed with a relatively small number of training points (few hundred and up to few thousand at most) in all benchmark examples. The closest studies to ours are \cite{brevis2020}
and \cite{roth2021neural}, where in the latter only the adjoint equation is approximated via a neural network. Using neural networks 
for both the primal and the adjoint equation is a difference to that prior works. Moreover, the extension to multiple goal functionals has not yet been treated in the state-of-the-art literature.
By embedding deep learning modules we have build a novel and reliable framework that has been tested and validated for both linear and non linear PDEs in complex domain, with various boundary conditions and numerous goal functionals.  \\
\\
The paper is structured as follows: In Sect.~\ref{sect1} we introduce the model problem and discuss about the neural network features and implementation details for training purpose. Next in Sect.~\ref{sect2} the basic concept of goal functionals and the approach for dealing  multiple goal functionals are presented. Sect.~\ref{sect3} briefly reviews the NURBS geometry and its application to represent complex geometry. In Sect.~\ref{sect4}, several
numerical tests substantiate our algorithm developments. Finally,
in Sect.~\ref{sec_conclusions}, conclusions are drawn and some ideas for are proposed future studies.

\section{Problem setup \& methodology}
\label{sect1}

Let us consider a parameterized \footnote{$\lambda$ is real, for example in the Navier-Stokes problem it represents the viscosity of the fluid, Reynolds number etc.} PDE in the following form:
\begin{align*}
  \mathcal{N}(u,\nabla u(\bf{x}),\lambda)=f(\bf{x})
\end{align*}
subjected to certain BCs (Dirichlet, Neumann or mixed) \i.e, for $\bf{x} \in \partial \Omega$
\begin{align*}
    \mathcal{B}[u(\bf{x})]= g(\bf{x})
\end{align*} where $\mathcal{N} : \mathcal{V} \mapsto \mathbb{R}$ is any differential operator (linear or non linear), $\bf{x} \in \Omega\subset \mathbb{R}^n$ and define,
\begin{align*}
    \Gamma(\bf{x},u) : = \mathcal{N}(u,\nabla u(\bf{x}),\lambda) - f(\bf{x})
\end{align*}
Here $\mathcal{V}$ is some suitably chosen function space containing the exact solution and the family of all neural network functions  corresponding each set of deep collocation points covering the domain $\Omega$.\\
This version of the boundary value problem (BVP) is called strong form. We proceed by approximating $u$ and $\Gamma$ using Deep Neural Networks (DNNs) $U_{\theta} = U(\bf{x},\theta)$ and $\Gamma_{\theta}= \Gamma(\bf{x},U_{\theta})$. Our approach adopts  deep collocation method, which assumes a certain discretization of the domain $\Omega$ and its boundary $\partial \Omega$ into a collection of points $\pi_{\Omega}$ and $\pi_{\partial \Omega}$ respectively. The goal is to learn the parameters of NNs. These along with  parameters of the operator $\mathcal{N}$ are learned by minimizing the error function under mean squared error (MSE) norm. For $\bold{x}_i \in \pi_{\Omega}$ and $\bold{s}_i \in \pi_{\partial \Omega}$ define :
\begin{align}
\label{eq1}
\mathcal{E}[\theta] : & = \frac{1}{N_{int}} \sum_{i=1}^{N_{int}}|\Gamma_{\theta}(\bold{x}_i)|^2 +  \frac{1}{N_{bnd}} \sum_{i=1}^{N_{bnd}} |\mathcal{B}[u(\bold{s}_i)]-g(\bold{s}_i)|^2  \nonumber \\ & = ||\Gamma_{\theta}(\bf{x})||_{\pi_{\Omega}}^2~~+ ||\mathcal{B}[u(\bold{s})]-g(\bold{s})||_{\pi_{\partial \Omega}}^2 
\end{align} 

The points are uniformly sampled from the domain and its boundary. Basically this equation is a Monte Carlo approximation of
    \[ Err[\theta] := \underset{x \sim U(\Omega)}{\mathbb{E}} \left[(\Gamma_{\theta}(\bf{x}))^2 \right] + \underset{s \sim U(\partial \Omega)}{\mathbb{E}} \left[(\mathcal{B}[u(\bold{s})]-g(\bf{s}))^2 \right] \]
where $U(\Omega)$ and $U(\partial\Omega)$ represent the uniform distribution on the domain and its boundary. This error function measures how well a neural network approximates the original solution and satisfies BVP w.r.t the defined norm. From an optimization point of view it could be more sophisticated to multiply the constraints with suitable penalty coefficients. The procedure seems to be a major bottleneck of the algorithm and if naively chosen minimization becomes more difficult that leads to a computationally expensive simulation. Readers are referred to~\cite{scheithauerjorge}  for more information about penalty methods.  In this contribution a characteristic feature used in the approach is that  error loss directly deals with the strong form, circumventing the necessity of  deriving the variational formulation. Moreover a prior knowledge of the exact solution is redundant here. Given any BVP the error loss can be computed directly up to a certain level of approximation of the solution. The main objective is to construct such a  neural network function $U_{\theta}$ by fine tuning the parameters so that $\mathcal{E}$ is as close to $0$ as possible.
\subsection{Network architecture}
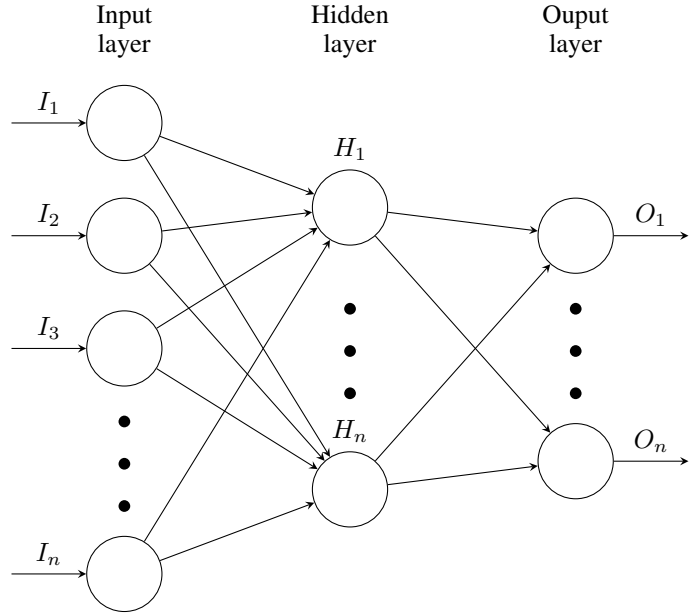
\begin{figure}[t]
	\centering
	\tikzset{%
  every neuron/.style={
    circle,
    draw,
    minimum size=1cm
  },
  neuron missing/.style={
    draw=none, 
    scale=4,
    text height=0.333cm,
    execute at begin node=\color{black}$\vdots$
  },
}

\begin{tikzpicture}[x=1.5cm, y=1.5cm, >=stealth]

\foreach \m/\l [count=\y] in {1,2,3,missing,4}
  \node [every neuron/.try, neuron \m/.try] (input-\m) at (0,2.5-\y) {};

\foreach \m [count=\y] in {1,missing,2}
  \node [every neuron/.try, neuron \m/.try ] (hidden-\m) at (2,2-\y*1.25) {};

\foreach \m [count=\y] in {1,missing,2}
  \node [every neuron/.try, neuron \m/.try ] (output-\m) at (4,1.5-\y) {};

\foreach \l [count=\i] in {1,2,3,n}
  \draw [<-] (input-\i) -- ++(-1,0)
    node [above, midway] {$I_\l$};

\foreach \l [count=\i] in {1,n}
  \node [above] at (hidden-\i.north) {$H_\l$};

\foreach \l [count=\i] in {1,n}
  \draw [->] (output-\i) -- ++(1,0)
    node [above, midway] {$O_\l$};

\foreach \i in {1,...,4}
  \foreach \j in {1,...,2}
    \draw [->] (input-\i) -- (hidden-\j);

\foreach \i in {1,...,2}
  \foreach \j in {1,...,2}
    \draw [->] (hidden-\i) -- (output-\j);

\foreach \l [count=\x from 0] in {Input, Hidden, Ouput}
  \node [align=center, above] at (\x*2,2) {\l \\ layer};

\end{tikzpicture}
	\caption[Network graph for a $(\mathrm{L}+1)$-layer perceptron.]{Network graph of a $(\mathrm{L}+1)$-layer perceptron with $D$ input units and $C$ output units. The $\mathrm{L}^{\text{th}}$ hidden layer contains $m^{(l)}$ hidden units.}
	\label{fig1}
\end{figure}

An Artificial Neural Network (ANN) is a human brain inspired algorithm and is designed to approximate a continuous mapping based on the given input.
The network chain categorized into three different classes of layers namely: input layer, hidden layers and output layer and each layer is made up of multiple neurons. The structure is shown in Fig.~(\ref{fig1}). Again neurons in each layer is connected to every other neurons in the next layer. For example two adjacent layers  are coupled as follows :
\begin{align*}
    y_{\ell}(\bm{x}) : = \sigma_{\ell}(\mathcal{W}_{\ell}y_{\ell-1}(\bm{x})+\beta_{\ell})~~~~;~~\ell=1,2,3\ldots,\mathrm{L}+1
\end{align*}
where, $y_{\ell}$ is the output of layer  $\ell$,   $\mathcal{W}_{\ell}$  are affine mappings and $\sigma_{\ell}$ is a fixed element wise activation function. The affine mappings are combinations of rectangular matrix whose elements are called weights, and $\beta_{\ell}$ is a bias vector.  Consequently using the functional structure of DNN as a sequence of function composition or layers the relationship between an input vector $\bold{x}_0$ and output prediction $\bold{y}_{\mathrm{L}+1}$ can be expressed in the form : 
\begin{align*}
  \bold{y}_{\mathrm{L}+1} = \mathfrak{N}(\bold{x}_0;\bold{\theta})= \mathcal{W}_{\mathrm{L}}  \sigma_{\mathrm{L}} \left(\ldots  \sigma_3 \left( \mathcal{W}_{2} \sigma_2 \left(\mathcal{W}_1 \bold{x}_0 +\bold{\beta}_1 \right) +\bold{\beta}_2 \right) \ldots \right) +\bold{\beta}_{\mathrm{L}}  
\end{align*}
where, $\bold{\theta}:= \left\{ \mathcal{W}_1, \bold{\beta}_1;  \mathcal{W}_2, \bold{\beta}_2; \ldots; \mathcal{W}_{\mathrm{L}} , \bold{\beta}_{\mathrm{L}}  \right\}$
is a set of all trainable parameters. Now suppose we have a sample of $T$ observation data $\{x_1,x_2,\ldots,x_T\}$ then training an ANN means to solve the optimization problem 
\begin{align*}
 \bold{\theta}_{\ast} : = \arg \min_{\bold{\theta} \in \Theta} \sum_{i=1}^T \left[ u(x_i) - \mathfrak{N} (x_i;\bold{\theta}) \right]^2   
\end{align*}
where $\Theta$ is a collection of all admissible parameters.\\
The optimized numbers of required neurons and layers are the main hyper-parameters controlling the architecture of this black box model. There is no analytical way to calculate or easy rule of thumbs to follow. Depending on the nature of problems optimal values needed to be explore in order to avoid over-fitting and under-fitting inside the network. The signals are transmitted through this connection from one layer to another, while each input is multiplied by a set of weights and  bias values associated with these, determines the parameter $\bold{\theta}$. The whole signals are now summed together to feed  the node's activation function. The role of activation function is to introduce non linearity into the network. This is one of the crucial component in the network and if fits well it can reduce the computational cost while retaining the efficiency and accuracy of the model.  

Once the architecture has been decided \i.e, the number of layers, number of neurons on each layer, and the activation functions are frozen, the problem boils down to determine the parameter $\bold{\theta}$. In our problem  the process of obtaining these parameters through training the network or equivalently the problem of minimizing the loss function~Eq~\ref{eq1}. Here comes the importance of optimizers algorithm that are used to update those parameters in order to reduce the losses and provide the most accurate results as possible. As like activation functions numerous optimizers exist in the literature but the challenge is to segregate the useful one for gaining the optimal solution sacrificing nominal computational cost. 


\subsection{Implementation}
The success of a neural network completely depends on its architecture. There is no rule of thumb for selecting, and therefore different problems ask for different architectures. Nevertheless a wise construction of an architecture can significantly improve the performances. Our novel architecture is build upon using \emph{Keras} model~\cite{Keras} with \emph{Tensorflow} backend. We don't use original Keras instead build Keras model with tf.keras module. Even though standard gradient descent optimizers is dominating the training process of DNN model these days, however the estimation is cruder and therefore  using a second order method for example L-BFGS  attempts to compensate and results in better generalizations. The optimizers are imported from \emph{Tensorflow Probability}~\cite{TP} library to train tf.keras.model and its subclass. One of the main advantage of this workaround is this library is GPU-capable and hence it is possible to perform large-scale computations.   Let us also briefly discuss about the activation function. As mentioned earlier it is the building blocks and requires to sort out between \enquote{useful} and \enquote{not so useful} from the plethora of information. Generally sigmoid or $\tanh$ are the most well known activation functions used in various deep learning models~\cite{saha2019evolution,samaniego2020energy} and we have also started with such functions but  fail to obtain the optimum results. Therefore in our experiments we have proceeded via trial and error methods to make a better choices of an activation function for easy and fast convergence of the network. One thing to note that the final layer should always be the linear function of its preceding layers.\\
We now provide the implementation details of the algorithm  and have highlighted some typical steps that one would have to follow for a Python based Tensorflow~\cite{abadi2016tensorflow}, an open source well documented  and currently one of the most popular, fastest growing deep learning library. The idea is to demonstrate the general procedure. The full source code is available on Github and can be obtained by contacting the authors.\\

By using the Xavier Initialization technique~\cite{glorot2010understanding}  the weights and biases are initialized in the following manner :

\begin{lstlisting}
  def initialize_NN(self,layers):
        weights = []
        biases = []
        num_layers = len(layers)
        for l in range(0, num_layers - 1):
            W = self.xavier_init(size=[layers[l], layers[l + 1]])
            b = tf.Variable(tf.zeros([1, layers[l + 1]], dtype=tf.float64), dtype=tf.float64)
            weights.append(W)
            biases.append(b)
        return weights, biases

    def xavier_init(self, size):
        in_dim = size[0]
        out_dim = size[1]
        xavier_stddev = np.sqrt(2.0 / (in_dim + out_dim))
        return tf.Variable(tf.truncated_normal([in_dim, out_dim], stddev=xavier_stddev), dtype=tf.float64)
\end{lstlisting}
$u(\bf{x})$ can be simply defined as follows :
\begin{lstlisting}
def net_u(self,x,y):

        X = tf.concat([x,y],1)

        u = self.neural_net(X,self.weights,self.biases)
\end{lstlisting}
 Finally, the neural network is defined using an activation function :
  \begin{lstlisting}
       def neural_net(self,X,weights,biases):
       
        num_layers = len(weights) + 1
        
        H = 2.0*(X - self.lb)/(self.ub - self.lb) - 1.0
        for l in range(0,num_layers-2):
            W = weights[l]
            b = biases[l]
            H = tf.tanh(tf.add(tf.matmul(H, W), b))
        W = weights[-1]
        b = biases[-1]
        Y = tf.add(tf.matmul(H, W), b)
        return Y
\end{lstlisting}

\section{Goal-oriented error estimation}
\label{sect2}

In this section, the main objective is goal-oriented error estimation using adjoint sensitivity weights. This idea was originally proposed in \cite{BeRa96,becker2001optimal} 
and is the so-called dual-weighted residual (DWR) method. 

\subsection{Main concepts and goal functionals}
In the following, we describe the DWR approach in more detail. Often, instead of an accurate approximation  $u_{\theta}$ everywhere in the computational domain, our focus of interest is a quantity of interest $J(u_{\theta})$, which should be computed to a certain accuracy. The objective is to measure the error $e$ between the (often unknown) exact solution $u$ and the 
(neural network) approximate solution $u_{\theta}$
in terms of a goal functional $J: \mathcal{V} \to \mathbb{R}$ :
\[
e = J(u) - J(u_{\theta}).
\]
Examples for such goal functionals are the mean values, line integration or point value evaluations:
\[
J(u) = \int_{\Omega} u \, dx, \quad J(u)= \int_{\partial \Omega} \frac{\partial u}{\partial n} \,ds, \quad J(u) = u(x_0) \quad\text{for } x_0\in \Omega.
\]
Moreover technical quantities such as stress evaluations, drag, lift, and 
vorticity evaluations can be represented with the help of $J(\cdot)$.

The second goal functional can be computed using Green's first identity. Following \cite{becker2001optimal}, the goal functional evaluation can be framed into an optimization 
problem in which the goal functional error (adjoint based) is minimized with respect to the governing 
partial differential equation. Hence
\[
\min (J(u) - J(u_{\theta}))  \quad\text{s.t.} \quad \mathcal{N}(u) = f.
\]

   The above goal functional error is computed with the numerical method leading to a discrete version $u_{\theta}$. However the main objective is to control the error $J(u) - J(u_{\theta})$ in terms of the local residuals computed in each (collocation) point. To address this we assign an associated adjoint problem, for which we use the Lagrangian formalism. 
   The Lagrangian $L:\mathcal{V}\times\mathcal{V}\to \mathbb{R}$ is defined as \[
   L(u,z) = (J(u) - J(u_{\theta})) - \mathcal{N}(u)z + fz,
   \]
   with the adjoint variable $z\in\mathcal{V}$.
   The corresponding optimality system is obtained by differentiation with respect to $u$ and $z$ resulting in the primal and adjoint equations: 

   Find $u , z \in \mathcal{V}$ such that
\begin{align}
\label{eq3}
\left.\begin{aligned}
    \mathcal{N}(u) - f &= 0 \\
    \mathcal{N}'(u)z - J'(u) &= 0
    \end{aligned} \qquad \right\}
\end{align}

where $\mathcal{N}'$ and $J'$ represent the Fr\'echet derivatives of the original operator $\mathcal{N}$ and goal functional $J$, respectively.
\begin{remark}
For linear PDEs $\mathcal{N}(u)$ and linear goal functionals 
$J(u)$, the above system reads
\begin{align}
\label{eq3_lin}
\left.\begin{aligned}
    \mathcal{N}(u) - f &= 0, \\
    \mathcal{N}(z) - J(u) &= 0.
    \end{aligned} \qquad \right\}
\end{align}
\end{remark}

In order to solve System \eqref{eq3}, we approximate $u$ and $z$ by 
their neural network approximations $u_{\theta}$ and $z_{\theta}$, 
respectively. Find $u_{\theta}$ and $z_{\theta}$ such that
\begin{align*}
    \mathcal{N}(u_{\theta}) - f &= 0 \\
    \mathcal{N}'(u_{\theta})z_{\theta} - J'(u_{\theta}) &= 0.
\end{align*}
We see that the adjoint problem yields 
as solution a sensitivity measure $z_{\theta}$ of the primal 
solution $u_{\theta}$ with respect to the given functional of 
interest $J'(u_{\theta})$.

Then, the main DWR theorem \cite{becker2001optimal} yields the error 
identity
\begin{align*}
    e = J(u)-J(u_{\theta}) = \frac{1}{2}\rho (u_{\theta})(z-z_{\theta})
    +\frac{1}{2}\rho^*(u_{\theta},z_{\theta})(u-u_{\theta}) 
    +  \mathcal{R}^{(3)} 
\end{align*}
with the primal $\rho$ and adjoint $\rho^*$ error parts
\begin{align*}
    \rho (u_{\theta})(\cdot) &\coloneqq -\mathcal{N}(u_{\theta})(\cdot) + f(\cdot), \\ 
    \rho^*(u_{\theta},z_{\theta})(\cdot) &\coloneqq J^{\prime}(u_{\theta})(\cdot) - \mathcal{N}^{\prime}(u_{\theta})(\cdot,z_{\theta}) +f(\cdot).
\end{align*}
For linear problems, we obtain immediately
\[
e = J(u)-J(u_{\theta}) = \rho (u_{\theta})(z-z_{\theta}).
\]
For mildly nonlinear (i.e., known as semi-linear in PDE analysis; equations in which the highest derivative is still linear) problems as for instance Navier-Stokes, still the 
primal error part $\rho (u_{\theta})(z-z_{\theta})$ is often used:
\begin{align}
\label{eq_primal_error}
    e = J(u)-J(u_{\theta}) = \rho (u_{\theta})(z-z_{\theta})
    +  \mathcal{R}^{(2)} 
\end{align}
with a quadratic remainder term. By approximating the difference 
$z-z_{\theta} =: \tilde{z}_{\theta}$ by a global higher order solution or local higher order 
interpolation and neglecting the remainder term, we obtain an approximation, which is used as computable error estimator $\eta$:
\begin{align}
\label{eq_primal_err_comp}
    e = J(u)-J(u_{\theta}) \approx \eta := \eta(u_{\theta},z_{\theta}) 
    := \rho (u_{\theta})(\tilde{z}_{\theta})
\end{align}
The approximation quality is measured with the help of the effectivity index
\[
I_{eff} := \frac{\eta}{e}.
\]
Asymptotically, we aim for $I_{eff} \to 1$. 

The approximations $u_{\theta}$ and $z_{\theta}$ are traditionally obtained 
with the help of finite elements \cite{hughes2012finite,Cia87} or isogeometric analysis \cite{cottrell2009isogeometric,nguyen2015isogeometric}.
In \cite{roth2021neural}, $u_{\theta}$ was computed with finite elements,
and $z_{\theta}$ was obtained by a neural network solution.

The main novelty in this paper is that we compute both 
$u_{\theta}$ and $z_{\theta}$ with the help
of neural networks as explained in the previous section.
Since in practice we are in general unaware about the exact adjoint solution $z$ inside $\tilde{z}_{\theta}$, we simply use in this work 
$\tilde{z}_{\theta} := z_{\theta}$. To this end, we have

\begin{prop}
Given the neural network approximated primal $u_{\theta}$ and adjoint $z_{\theta}$ solutions for the boundary value problem 
$\mathcal{N}(u) = f$ and the goal functional $J(u)$, we have 
as error estimator using the primal error part \eqref{eq_primal_err_comp}
\begin{align*}
    \eta =
    \rho (u_{\theta})({z}_{\theta}) = f({z}_{\theta}) -\mathcal{N}(u_{\theta})({z}_{\theta})
\end{align*}
and as a posteriori error estimate
\begin{align*}
    |e| = |J(u)-J(u_{\theta})| \leq |\eta| =
    |f({z}_{\theta}) -\mathcal{N}(u_{\theta})({z}_{\theta})|.
\end{align*}
Both residual terms are assembled pointwise by summing up such that
\begin{align*}
f(z_{\theta}) &=\frac{1}{N}\sum_{i=1}^{N} f(z_{\theta})_i, \\
\mathcal{N}(u_{\theta})(z_{\theta}) & = \frac{1}{N}\sum_{i=1}^{N} \mathcal{N}(u_{\theta_i}) \cdot z_{\theta_i}
\end{align*}
with $f(z_{\theta})_i := f_i \cdot z_{\theta_i} $ in each collocation point and for $N=N_{int}+N_{bnd}$ and where $z_{\theta_i}$ is the adjoint solution (obtained from the neural network) in each collocation point and summing over $i$ means that we evaluate on each collocation point in the domain.

\end{prop}

\subsection{Multiple goal functionals}
\label{ss1}
The main purpose of this section is to extend the previous approach from 
single to multiple goal functionals. In many applications (specifically 
multiphysics problems) several quantities of interest are simultaneously 
of importance.
Given $\aleph$ target functionals say $J_1,J_2,\ldots,J_{\aleph}$ the standard procedure to compute the error estimators analogous to~Eq.(\ref{eq3}), \i.e, for each individual $J_n$, we introduce an adjoint problem: Find $z_n$ such that
\begin{align*}
 \mathcal{N}'(u)(z_n) - J_n'(u) = 0, \quad n=1,2,\ldots,\aleph.  
\end{align*}
Computing each adjoint problem separately involves the expensive cost of $\aleph$ computations. However this can be avoided following the techniques proposed~\cite{hartmann2003goal,hartmann2008multitarget} with 
further improvements and extensions proposed in \cite{endtmayer2017partition,endtmayer2019multigoal} (see also \cite{Wi20_book,Endt21}) of combining the functionals. The core idea is to construct a linear combination to form a \emph{combined target functional}
\begin{align*}
    J_c(\cdot)= \sum_{n=1}^{\aleph} \omega_nJ_n(\cdot)
\end{align*}
where,
\begin{align}
\label{g1}
    \omega_n:= \frac{\text{sign}\{J_n(u)-J_n(u_{\theta})\} w_n}{|J_n(u_{\theta})|} 
    = \mu_n w_n
\end{align}
and $w_n$ are self chosen positive weights. Such choices lead to no error cancellation, however depending on the weights any particular functional may impact on $J_c$. We may now proceed similarly as in single functionals to derive an error representation formula for the errors in $J_c$:

\begin{prop}
Given the neural network approximated primal $u_{\theta}$ solution
for the (discrete) boundary value problem 
$\mathcal{N}(u_{\theta}) = f$ and adjoint  solution $z_{\theta}$ obtained by
$
 \mathcal{N}'(u_{\theta})(z_{\theta}) = J_c'(u_\theta),
$
with the combined goal functional $J_c$, we have as error estimator
using the primal error part
\begin{align*}
    \eta =
    \rho (u_{\theta})({z}_{\theta}) = f({z}_{\theta}) -\mathcal{N}(u_{\theta})({z}_{\theta})
\end{align*}
and as a posteriori error estimate
\begin{align*}
    |e| = |J_c(u)-J_c(u_{\theta})| \leq |\eta| =
    |f({z}_{\theta}) -\mathcal{N}(u_{\theta})({z}_{\theta})|.
\end{align*}
\end{prop}

\subsection{Algorithm}
Our aim of the computation is to approximate the target 
quantity $J(\cdot)$ so that it holds $|J(u)-J(u_{\theta})| < TOL$ with some user defined $Tol>0$. In the case of multiple goal functionals, we aim 
for $|J_c(u)-J_c(u_{\theta})| < TOL$.
The whole algorithm is tailored for an accurate and efficient approximation 
in order to obtain $I_{eff} \approx 1$. In practice the stopping criteria is enforced by a preassigned tolerance and if the condition is not satisfied then this can serve as an error indicator which is employed to guide the parameters in the neural network architecture. The weights and bias terms get updated accordingly and this process continues iteratively.\\
Summarizing all previous steps, we obtain as basic scheme:
\begin{algorithm2e}[htb]
\DontPrintSemicolon
\BlankLine
\textbf{Initialize :} Provide number of iteration steps, epochs and the neural network architecture.\;
\BlankLine
Generate collocation points $(x_i,y_i)$ on $\Omega$ \;
\BlankLine
Initialize the neural network $\mathfrak{N}(\cdot,\cdot)$\;
Initialize the weights of the network using Xavier initialization technique\; 
Obtain the loss function $\mathcal{E}[\theta]$ for the current mini-batch points\;

\BlankLine
{
 Minimize the loss\;
 \If {Loss $<$ \emph{TOL}} 
 {
   STOP\;
   }
     \Else { 
      repeat steps until the criteria is satisfied \;
     }
 \BlankLine    
Compute the adjoint solution $z_{\theta}$\;
Having just computed $u_{\theta}$ and $z_{\theta}$ construct error estimator $\eta(u_{\theta}, z_{\theta})$\;

  Evaluate $\eta(u_{\theta}, z_{\theta})$ and compare with true error $e$ \;
  Evaluate the quality of $\eta$ by observing the effectivity index $I_{eff}$\;
}
 \BlankLine
 \If { $I_{eff} \approx 1$} 
 {
   STOP\;
   }
     \Else { 
     adjust \emph{Tol} \& repeat steps until the criteria is satisfied \;
     }
 \BlankLine
\caption{Neural Network based goal oriented error approach}
\label{algo-synthesis}
\end{algorithm2e}

Then, using finer neural networks (in order to obtain a higher approximation quality) for computing $u_{\theta}$ and $z_{\theta}$ and repeating all previous steps, we can compare the different $I_{eff}$.

\section{Modelling based on NURBS curves}
\label{sect3}

Geometrical modelling of the computational domain is an essential tool for an accurate and computationally efficient approximation to the solutions of PDEs. Construction of a mesh based geometry is expensive and likely to creates inaccuracies for complex geometries. While this may be trivial for a regular shaped domain, but challenges arises when the domain is irregular for example \enquote{semi-annulus domain}. In this section we will briefly describe how NURBS based modelling are more effective and possesses the advantage of yielding highly accurate approximations. For a more detailed descriptions and applications see some classic literature~\cite{hughes2005isogeometric,cottrell2009isogeometric,piegl1996nurbs,nguyen2015isogeometric,dimas19993d,bingol2019nurbs} and the references therein.

\begin{figure}[htbp]
\centering
\includegraphics[width=3in,height=2in]{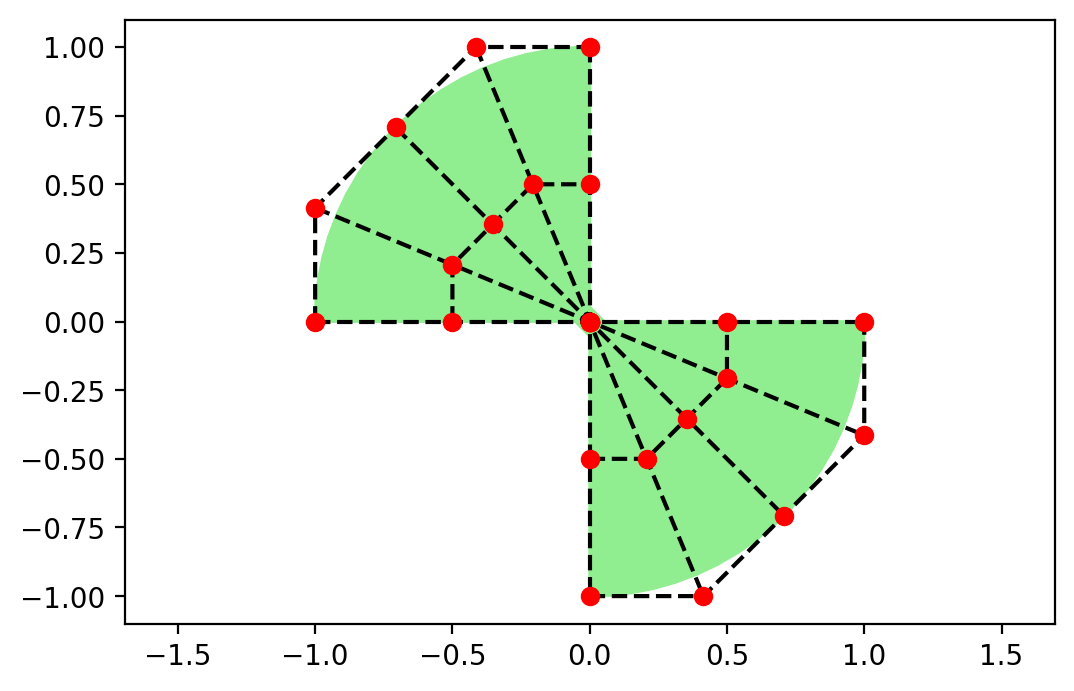}\quad 
\includegraphics[width=3in,height=2in]{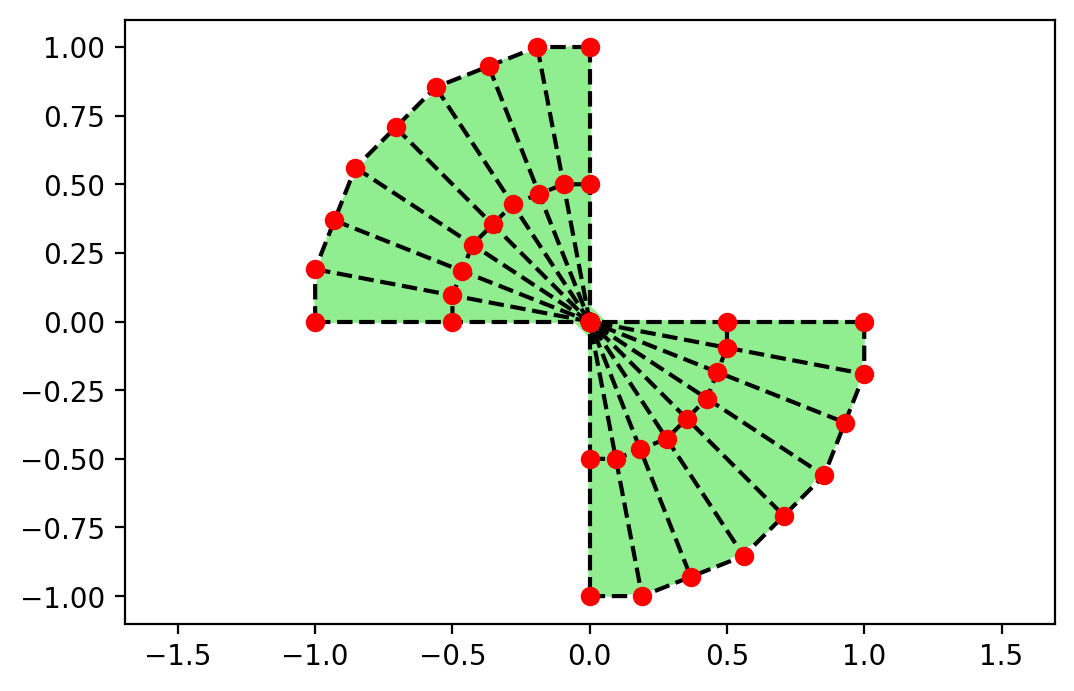}
\caption{(From left to right) Impact of knot vectors and control points on the shape of a closed curve.}
\label{fig6}
\end{figure}

NURBS (Non Uniform Rational B-Splines) are built from B-Splines. The basic B-Splines are lack of ability to exactly represent some simple shapes (circles, ellipsoids etc) and thus NURBS are the generalizations of B-Splines through rational functions so as to remove this difficulty. The geometric modeling typically consist of two meshes physical mesh and control mesh. While the physical mesh is a representation of the original geometry, the control mesh forms a scaffold of the geometry. NURBS shapes are defined by its degrees, knot vector and  set of control points which are the inputs supposed to be provided by the users. A knot vector is an increasing set of coordinates in the parametric space generally denoted by $\Xi=\{\xi_k \}_{k=1}^{n+p+1}$ where $\xi_k$ is the $k$-th knot, $n$ is the number of basis functions and $p$ is the polynomial order. Depending on the degree of basis functions and  the no of control points a knot can be repeated several times and it is even possible to insert a new knot without even changing the curve geometry and parameterization see Fig.~(\ref{fig6}). Analogously this is known as $h$-refinement.  Knot refinement offers a wide range of tools to design and analyse the shape information. A univariate rational basis function is defined as : 

\begin{align*}
   \mathcal{R}_{k,p}(\xi) : = \frac{w_k N_{k,p}(\xi)}{\sum_{k=1}^n w_i N_{k,p}(\xi)} ~~;~~k\in[1,p+1]
\end{align*}

Here, $\{N_{k,p}(\xi)\}_{k=1}^n$ is  the set of basis functions of the B-Spline curves, $\{w_k~:~w_k>0\}_1^n$ is the set of NURBS weight. Choosing appropriate weights permits the definition of varieties of curves and in particular if all the weights are equal its reduces to the B-Spline basis. These basis functions are multiplied with a set of weights and control points and summed up to generate a NURBS curve. Correct shape definition of a curve is strictly connected with the proper selection of control points, weights and knots. The implementation is done using NURBS-Python (geomdl) library. Following snippet of code is an illustration how to generate a 2D NURBS curve and visualizing it using NURBS-Python.

\begin{lstlisting}
  from geomdl import NURBS
 
 
 class Disc(NURBS):
    """
    Class for defining a circular domain 
    Input: rad,center - radius and center of the circle
    """
    def __init__(self, rad, center):
        geomData = dict()
        
        # Users set degrees
        geomData['degree_u'] = x_1
        geomData['degree_v'] = x_2
        
        # Users set control points
        geomData['ctrlpts_size_u'] = y_1
        geomData['ctrlpts_size_v'] = y_2
                
        geomData['ctrlpts'] = [...,...,...,.....]
        
        geomData['weights'] = [...,...,...,.....]
        
        # Users set knot vectors
        geomData['knotvector_u'] = [a_1,a_2,..,....]
        geomData['knotvector_v'] = [b_1,b_2,..,....]
        super().__init__(geomData)
     
     # Plot the curve
     Disc.PlotSurf()
 \end{lstlisting}

\section{Numerical examples}
\label{sect4}
 In this section, we investigate adaptive refinement and error control using the DCM for different PDEs on complex domains. Specifically, we consider 
 the reguralized $p$-Laplacian, Poisson's problem, and the stationary incompressible Navier-Stokes equations resulting in total in five numerical tests.  The details of the network architecture such as relevant optimizers, activation functions, number of layers and neurons on each layers are provided in each example separately. To demonstrate the effectiveness of our method, we computed the reference solution, wherever mentioned on the same testing set and continued to investigate how well our error estimators perform in approximating the error by invoking the effectivity indices and also with various choices of weight parameters that lead to different levels of difficulties with respect to the problems. As mentioned earlier the implementation has been carried out in the TensorFlow framework. All of the following results obtained using Python version $3.8.5$ and TensorFlow version $2.5.0$.  Unless mentioned otherwise  the numerical examples reported were run on a $64$-bit LinuxOS with Intel(R) Core(TM) i$7$-$8850$H CPU, $64$GB memory. Nevertheless training on a CPU is significantly slower than training it on a GPU, which is why the training part was also performed on a GPU whenever required.

\subsection{Example - $1$: Regularized $p$-Laplacian}

\begin{figure}[htbp]
\centering
\includegraphics[width=3.1in,height=2.3in]{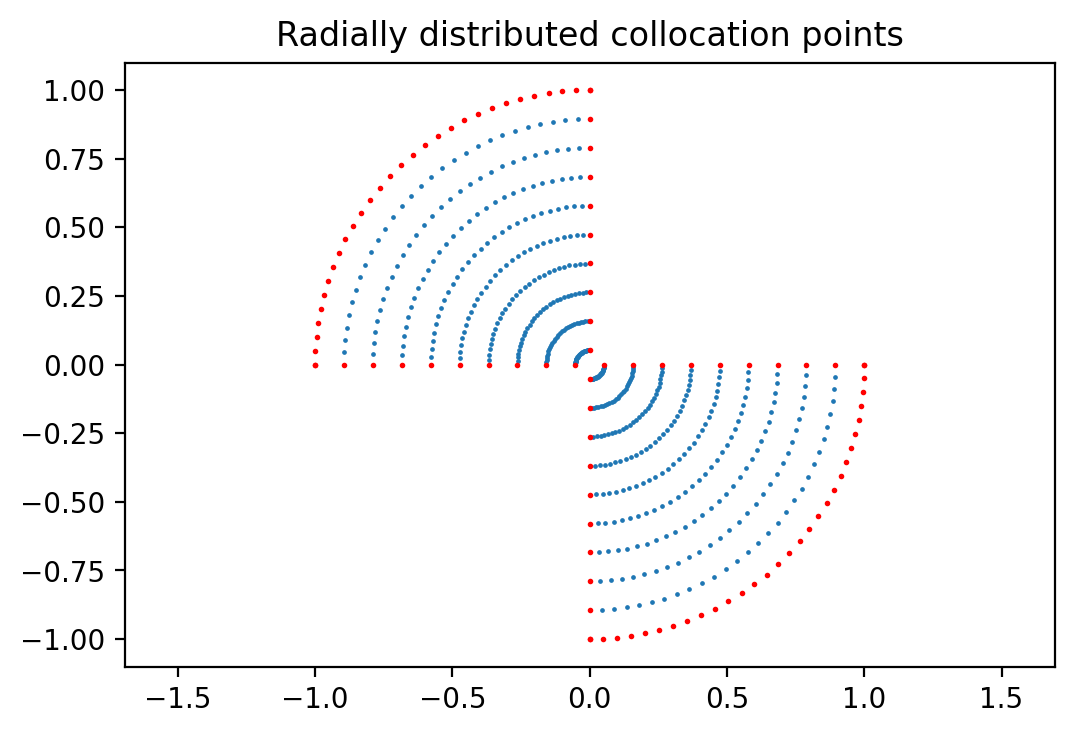}\quad 
\includegraphics[width=3.1in,height=2.3in]{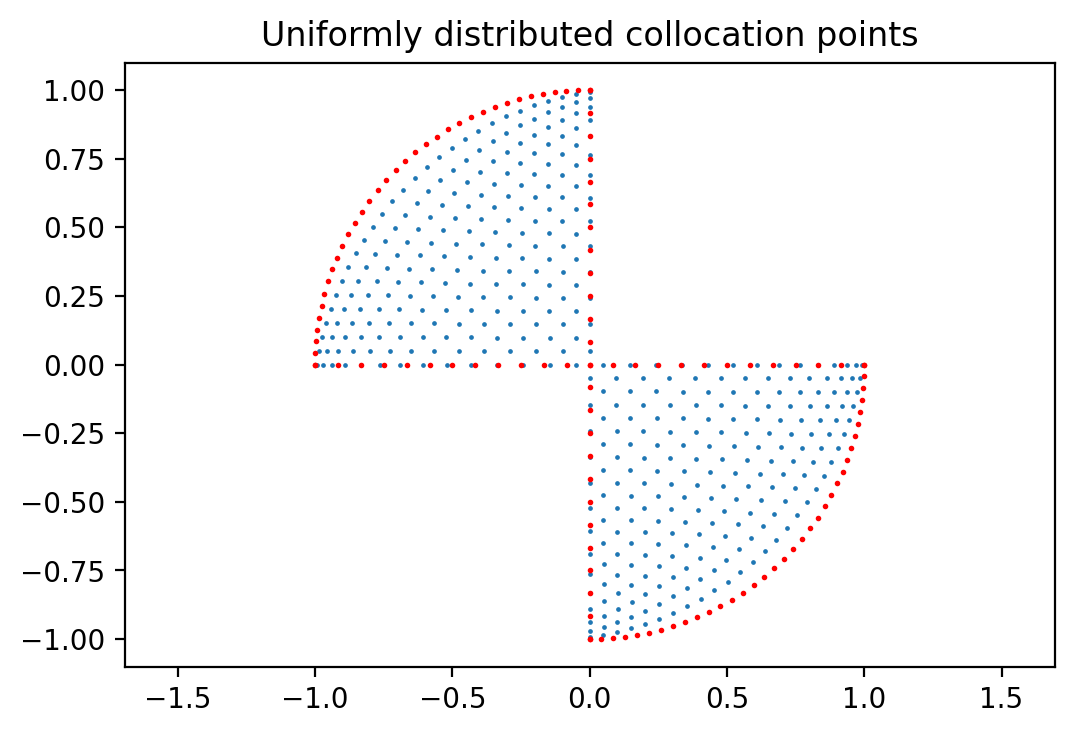} \\
\includegraphics[width=3.1in,height=2.3in]{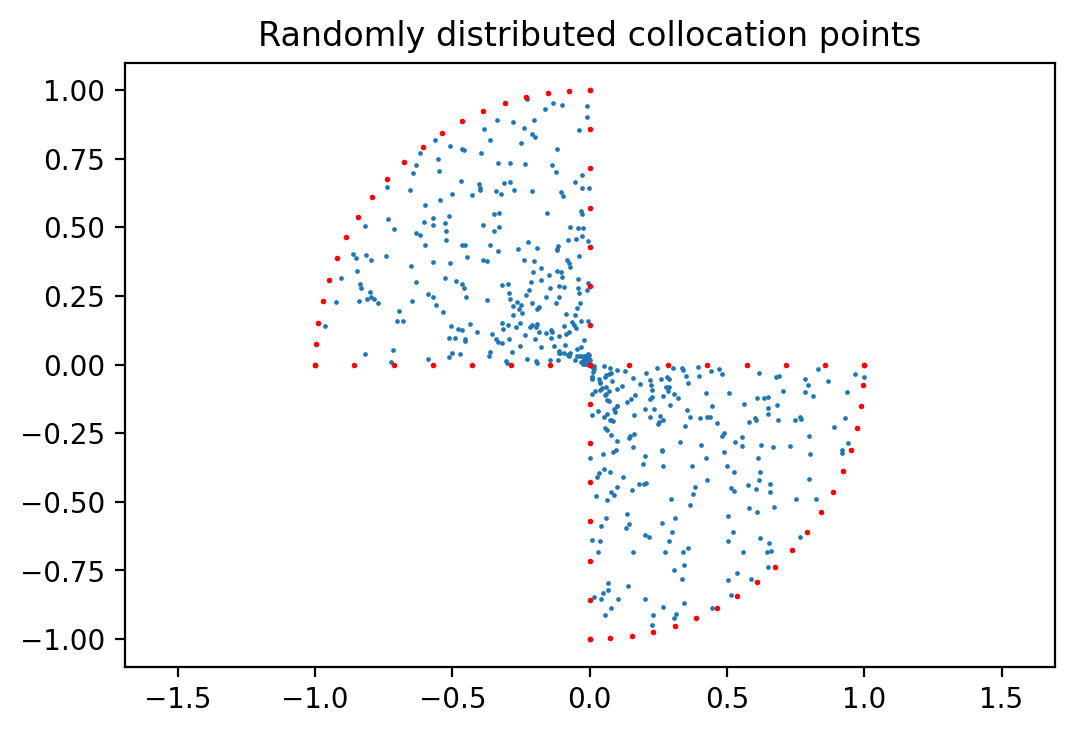} \quad
\includegraphics[width=3.1in,height=2.27in]{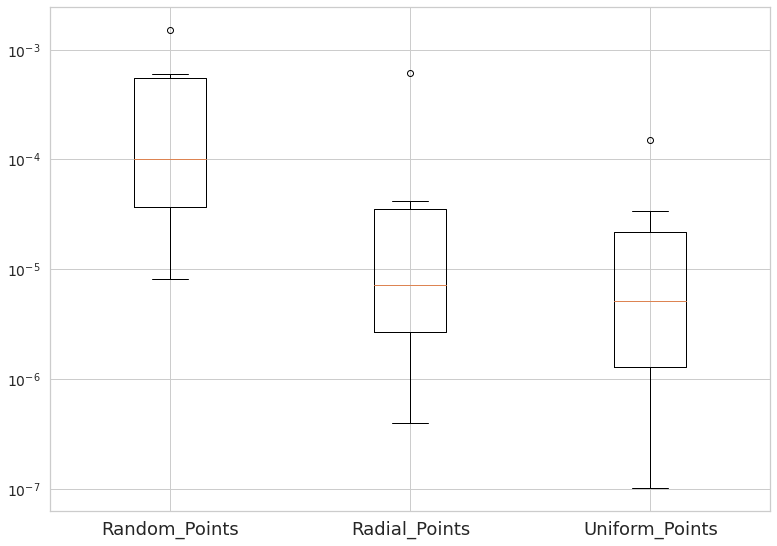}%
\caption{(From top left)Boundary and Interior collocation points on $\Omega$. (Bottom right) Relative error in box plot. } \label{fig2}
\end{figure}

On a bounded Lipschitz domain $\Omega$ in $\mathbb{R}^2$, and for  $p\ge 2$ and $\delta \ge0$, we define
\begin{align*}
  \mathcal{A}_{\delta,p}(u) \equiv \textrm{div}((\delta^2+|\nabla u |^2)^{\frac{p-2}{2}}\nabla u)  
\end{align*}

We consider the family: Find $u$ such that,
\begin{align*}
    -\mathcal{A}_{\delta,p}(u) & = f~~~~\textrm{on } \Omega\\
    \mathcal{B}[u] & = g~~~~\textrm{on } \partial \Omega
\end{align*}
The Fr\'echet derivative $\mathcal{A}_{\delta,p}'(u)$ at the linearization point $u$ of the non linear operator can be found in~\cite{endtmayer2019multigoal}.
More specifically, we define as domain
\begin{align*}
  \Omega=\{(x,y)~:~0\le x^2+y^2 \le 1 ~\&~xy\le0 \}  
\end{align*}
This $2$D shape has been constructed by repositioning control points, modifying knot vectors , adjusting weights and geometric degrees of a fitted NURBS disc.

We consider the following cases. 
\subsubsection{Case - I $\left (p=2,\delta=0 \right)$} The first case considers the Poisson problem subjected to Dirichlet boundary conditions.  Let,
\begin{align}
&\left.\begin{array}{@{}l}
  - \mathcal{A}_{0,2}(u)  = 12 xy~~~~\textrm{on } \Omega\\
     \qquad u  = 0 \qquad \textrm{on } \partial \Omega
    \end{array}\right\} 
     \label{eq2}
\end{align}
and the exact solution of (\ref{eq2}) will be used as a reference solution which is given by $\mathcal{U}(x,y)= xy(1-x^2-y^2)$. The geometrical set up and the arrangement of collocation points are demonstrated in Fig.~(\ref{fig2}). It clearly depicts the influence of the arrangement of points. The neural networks are trained to approximate the solution $U_{\theta}$ which is obtained by the minimization of the loss function. The problem statement can be written as : 
\begin{align*}
 \mathcal{E}:=||\mathcal{A}_{\delta,p}(U_{\theta}) -f(x,y)||^2_{\Omega}+||U_{\theta}||^2_{\partial \Omega} 
 \label{eq3}
\end{align*}

The quantity of interest for this example is as follows :
\subsubsection*{Goal Functionals of Interest}
We consider the following functionals:
\begin{align*}
    J_{\partial}(u)&:= \int_{\Gamma} \frac{\partial u}{\partial n} \,ds \\
     J_{\Omega}(u)&:= \int_{\Omega} u(x,y) \,d(x,y) \\
     J_{\chi}(u)&:= \int_{\Omega} \chi(x,y) u(x,y) \,d(x,y)
\end{align*}
where,
$\chi(x,y) = 
\left\{
	\begin{array}{ll}
		y-x  & \mbox{if } x<y \\
		0  & \mbox{ otherwise } 
	\end{array}
\right.
$
We are interested in the following goal functionals :
\begin{align*}
  J_1(u):= J_{\partial}(u) & J_{\chi}(u) \qquad  J_2(u):=J_{\Omega}(u) J_{\chi}(u) \qquad J_3(u):= J_{\Omega}(u) J_{\partial}(u) \\
  \\
  & J_c(u):= 0.4 \mu_1 J_1(u)+ 0.01 \mu_2 J_2(u)+ \mu_3 J_3(u)
\end{align*}

\begin{figure}[htbp]
\centering
\includegraphics[width=3.1in,height=2.3in]{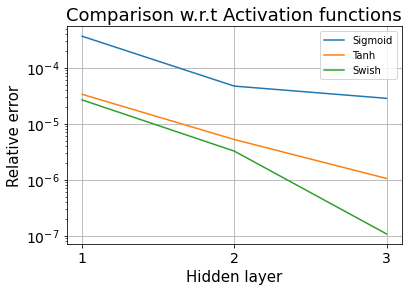}\quad 
\includegraphics[width=3.1in,height=2.3in]{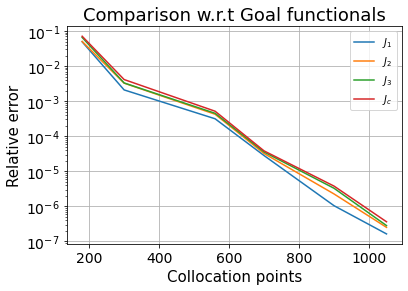} \\
\includegraphics[width=3.1in,height=2.3in]{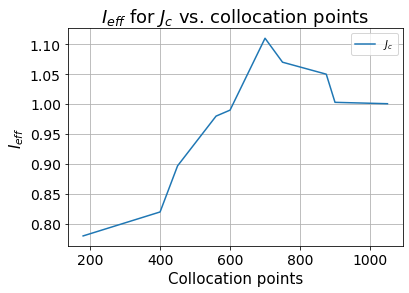} \quad
\includegraphics[width=3.1in,height=2.27in]{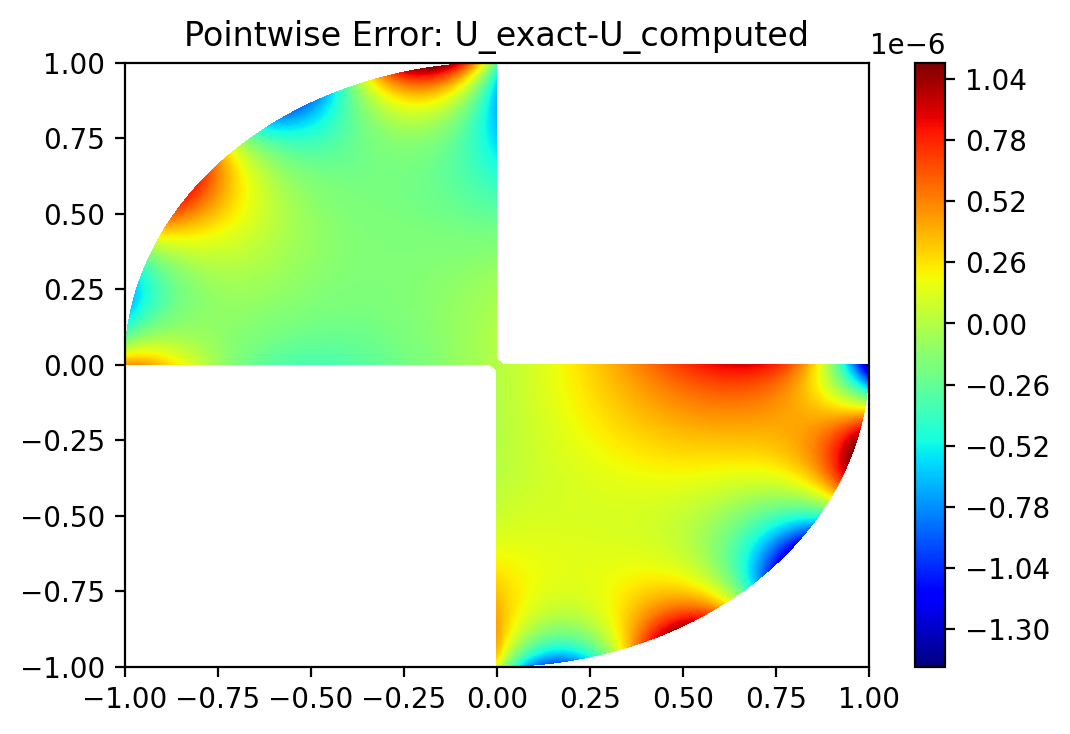}%
\caption{(left from the top) Comparisons of different Activation functions and Goal functionals w.r.t hidden layers and training points. (left from the bottom) Computed  effectivity index $J_c$ of the error estimator and error plots on the domain for the primal problem.}
\label{fig5}
\end{figure}

We have used a $5$ layers network $(2-30-30-20-1)$ and \emph{swish} activation function throughout the hidden layers to evaluate neural values in those layers. As an optimizer we used \emph{Adam} with learning rate $\alpha=0.001$ followed by \emph{BFGS}. The entire domain is discretized with $N_x \times N_y$ uniformly spaced training points where $N_x=35~,~N_y=30$. The training iterations has been stopped once the loss function reaches the tolerance~\emph{Tol}.=$10^{-12}$. In order to quantify the accuracy of our approach we compute the relative $\mathcal{L}_2$ error as:
\begin{align*}
\mathcal{L}_2 := \frac{\sqrt{\sum_{i=1}^{\mathcal{N}_{\text{pred}}}\mathcal{U}^2_{\text{err}}(x_i,y_i)}}{\sqrt{\sum_{i=1}^{\mathcal{N}_{\text{pred}}}\mathcal{U}^2(x_i,y_i)}}    
\end{align*}
where, $\mathcal{N}_{\text{pred}}$ is the no. of testing points and $\mathcal{U}_{\text{err}}(x,y)= \mathcal{U}(x,y)-U_{\theta}(x,y)$.\\
\\
Let us now discuss on our findings. From the observation in Fig.~(\ref{fig5}) we deduce that \enquote{swish} activation function outperforming in deep neural networks in comparison to other mostly used activation functions in the literature. Considering accuracy in the approximation of adjoint solutions of 
\begin{align*}
- \mathcal{A}'_{0,2}(u)  = J(u)~~~~\textrm{on } \Omega    
\end{align*}
we define reference solutions $\mathcal{U}_{adj}=c_J(1-x^2-y^2)$ and compare relative errors with respect to proposed goal functionals $J$. Here $c_J \in \mathbb{R}$ and its value needed to be adjusted according to  chosen $J$.  We observe that initially relative error in $J_c$ badly approximates the functional with the largest error, however it is decreasing steeply w.r.t the coll. points, \i.e, increasing collocation points deliver promising results. Similar behavior being also observed for $J_3$ because of the dominating weight $w_3$. Next we take a look at the effectivity indices for $J_c$ which are $I_{eff} \approx 1$. Therefore we conclude that we have almost approximate the true error with our error estimator. In particular the figure also shows that even with less amount of training points it is  possible to obtain a reasonable approximations of the true error. Finally the error plots in general reveals a good agreement between DCM solution and the reference solution. However there has some relatively larger difference in the vicinity of boundaries and it is because less no of training points here in comparison to the interior which is again a trade-off.

In this particular example we have generated around $1500$ training points ($N_x=42$ , $N_y=35$) over the entire domain as the feeding data to train our network. Tolerance used as $10^{-12}$. A $6$ layers neural network $(2-30-30-30-20-1)$ is constructed. Therein, we use $2$ neurons in the input layer and
$1$ neuron in the output layer. In addition $3$ hidden layers with $30$  neurons and $1$ hidden layer with $20$ neurons are used to enforce the learning behavior. Fig.~(\ref{fig8}) perfectly depicts the benefits of using deep neural network instead of shallow wide neural network. We use \enquote{swish} activation in the first three hidden layers and \enquote{tanh} in the final hidden layer to evaluate neural values in those layers. The learning rate $\alpha = 0.001$ is used in the \emph{BFGS} optimizer following \emph{Adam}.

\subsubsection{Case - II~~ $\left(p=2.5,\delta=0.05\right)$ }
In the second example on the same domain we consider following problem with mixed boundary conditions: Find $u$ such that,
\begin{align*}
     -\mathcal{A}&_{\delta,p}(u) = \frac{4\delta^2+20(x^2+y^2)}{\left[\delta^2+4(x^2+y^2) \right]^{0.75}}~~~~\textrm{on }\Omega\\
     & \left.\begin{array}{@{}l}
    u = 0~~~~\textrm{whenever, } x^2+y^2=1 ~\equiv \Gamma_1\\
    \frac{\partial u}{\partial x} = - 2x ~~~~\textrm{at, } y=0 ~\equiv \Gamma_2 \\
     \frac{\partial u}{\partial y} = - 2y ~~~~\textrm{at, } x=0~\equiv \Gamma_3
     \end{array}\right\} 
    \quad (x,y) \in \Gamma
\end{align*}

The exact solution is given by: $\mathcal{U}(x,y):= 1-x^2-y^2$. Similar to earlier example we intend to minimize the loss function with respect to the training data.

\begin{align*}
 \mathcal{E}:=||\mathcal{A}_{\delta,p}(U_{\theta}) -f(x,y)||^2_{\Omega}+ \left|\left| U_{\theta} \right|\right|^2_{\Gamma_1}+\left|\left|\frac{\partial U_{\theta}}{\partial x}- \mathcal{U}_x \right |\right|^2_{\Gamma_2} +\left|\left|\frac{\partial U_{\theta}}{\partial y}- \mathcal{U}_y \right |\right|^2_{\Gamma_3} 
 \end{align*}

\subsubsection*{Goal Functionals of Interest}
We are interested in the  following functionals evaluation:
\begin{align*}
    J_1(u):= & u(0.5,-0.5), \qquad 
    J_2(u):= \int_{\Omega} u(x,y) \,d(x,y), \qquad 
    J_3(u):= \int_{\Gamma} \frac{\partial u}{\partial n} \,ds
     \end{align*}
     and the combined functional
     \[
     J_c(u):= 0.5\mu_1J_1(u)+0.25 \mu_2 J_2(u)+\mu_3J_3(u).
    \]
In this particular example we have generated around $1500$ training points ($N_x=42$ , $N_y=35$) over the entire domain as the feeding data to train our network. Tolerance used as $10^{-12}$. A $6$ layers neural network $(2-30-30-30-20-1)$ is constructed. Therein, we use $2$ neurons in the input layer and
$1$ neuron in the output layer. In addition $3$ hidden layers with $30$  neurons and $1$ hidden layer with $20$ neurons are used to enforce the learning behavior. Fig.~(\ref{fig8}) perfectly depicts the benefits of using deep neural network instead of shallow wide neural network. We use \enquote{swish} activation in the first three hidden layers and \enquote{tanh} in the final hidden layer to evaluate neural values in those layers. The learning rate $\alpha = 0.001$ is used in the \emph{BFGS} optimizer following \emph{Adam}.
\begin{figure}[htbp]
\centering
\mbox{\subfigure{\includegraphics[width=3in,
  height=2.27in]{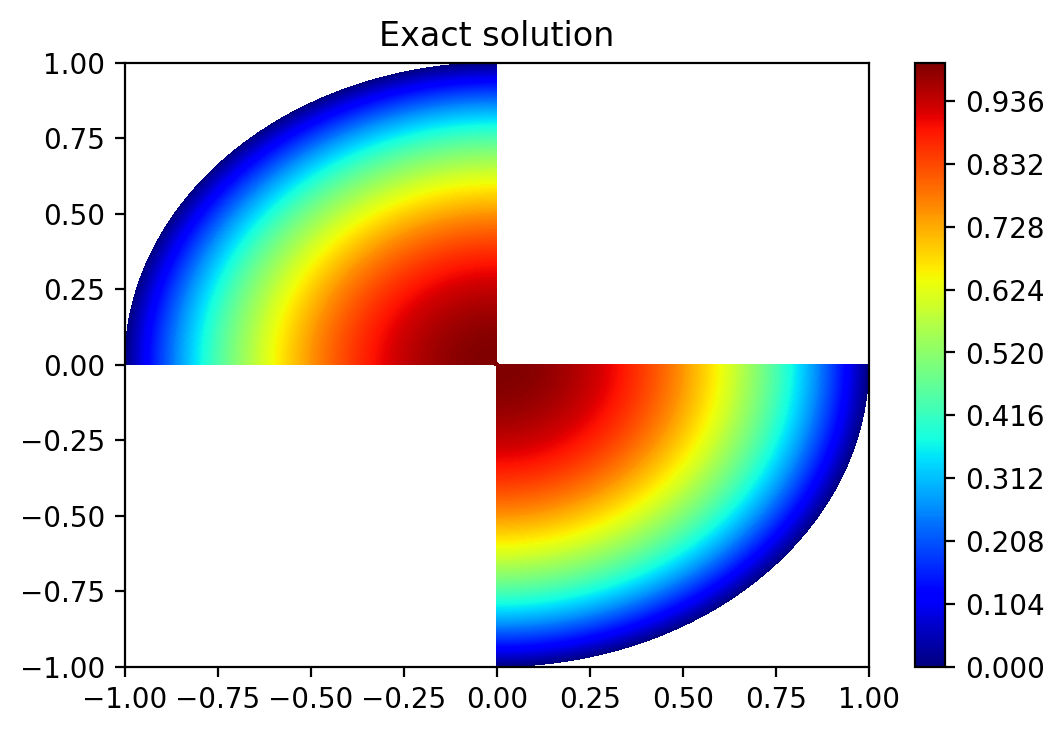}}\quad
\subfigure{\includegraphics[width=3in,height=2.27in]{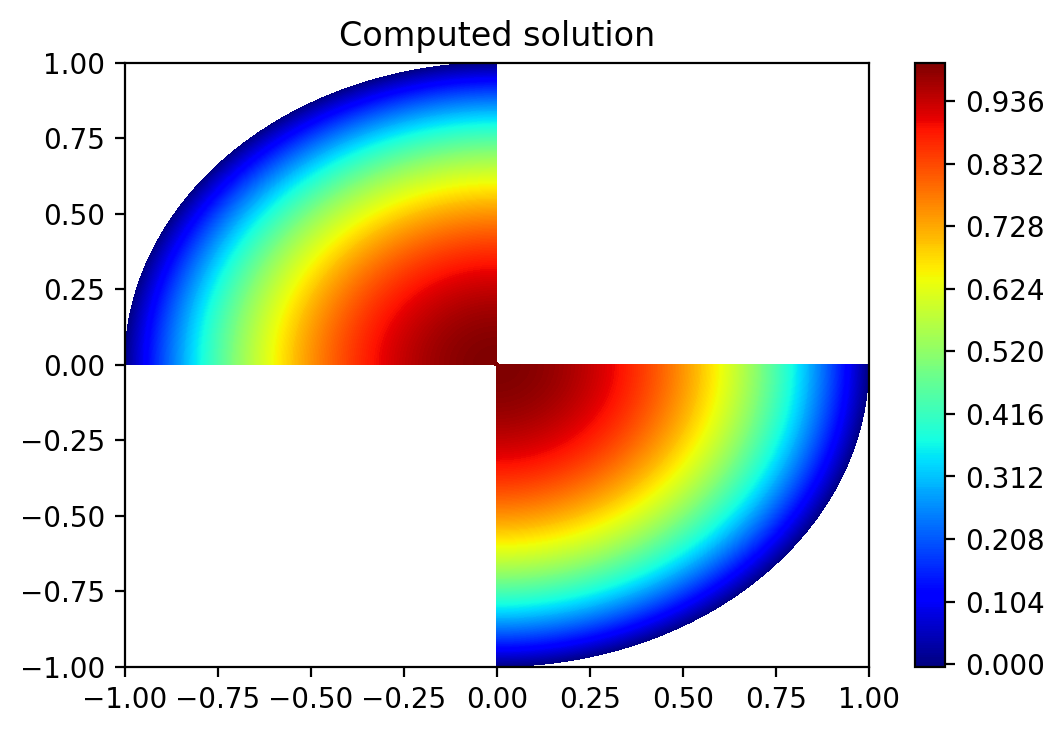} }}
\caption{Comparison of solutions for the primal problem} \label{fig3}
\end{figure}

\begin{figure}[htbp]
\centering
\mbox{\subfigure{\includegraphics[width=3.2in,
  height=2.5in]{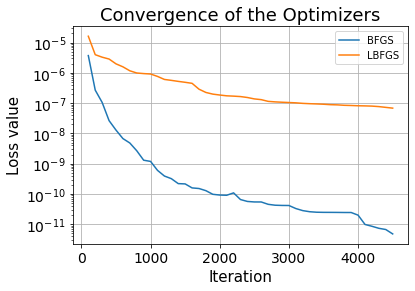}
  }\quad
\subfigure{\includegraphics[width=3.2in,height=2.5in]{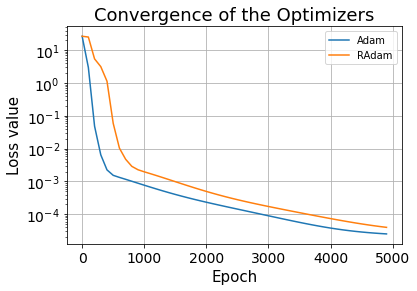} }}
\caption{Comparisons in convergence}
 \label{fig4}
\end{figure}
We don't know the exact solution for the adjoint problem in this case. Therefore in order to measure the accuracy we study the relative errors for the primal problem and residual errors for the adjoint problem. More hidden layers, with more neurons yield flatting of the  errors.  The residual error $\mathcal{R}_{\text{err}}$  is computed using the formula :
  \begin{align*}
    \mathcal{R}_{\text{err}}: =\left| \left|\mathcal{A}'_{\delta,p}(u) - J(u) \right| \right|^2_2
  \end{align*}

\begin{figure}[htbp]
\centering
\includegraphics[width=3.1in,height=2.3in]{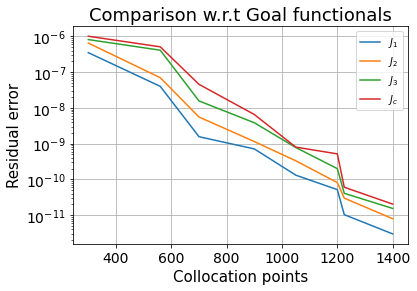}\qquad
\includegraphics[width=3.1in,height=2.3in]{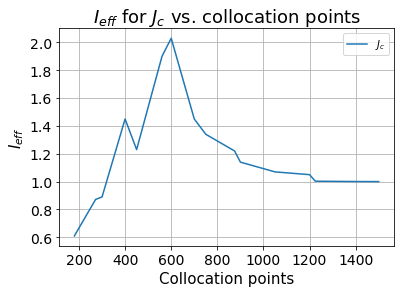}
\caption{Residual errors and effectivity index.}
\label{fig8}
\end{figure}  

\begin{figure}[htbp]
\centering
\includegraphics[width=3.1in,height=2.3in]{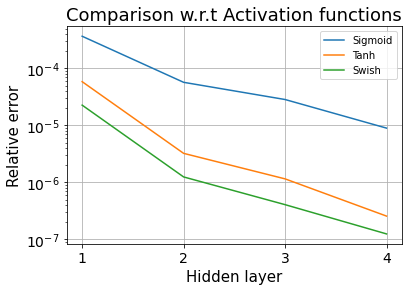}\qquad
\includegraphics[width=3.1in,height=2.3in]{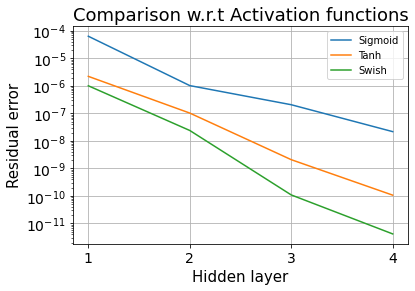}
\caption{Error vs. Hidden layers}
\label{fig9}
\end{figure}

 Considering the accuracy of the functional evaluations above, we observe that the residual errors in the functionals $J_1$, $J_2$, $J_3$ and $J_c$ are less than $10^{-10}$~cf Fig.~(\ref{fig8}). Furthermore, we observe that $J_c$ is the dominating functional and $J_1$ is the one with  least error. We noticed that the error in $J_3$ is similar to $J_c$. This is because $J_3$ is again the dominating part in the combination of $J_c$. Hence $J_c$ changes its behavior after the error of $J_3$ starts to dominate.  Therefore, we compute the effectivity index of this functional and obtain an excellent convergence towards $1$ as the no of collocation points start to increases from $1000$. As like before the interesting part is our methodology successfully approximate the true error with the error estimator.\\
 As a part of the network architecture depicted in Fig.~(\ref{fig4}) we conclude that (\emph{Adam},\emph{BFGS})  performs better than (\emph{RAdam},\emph{LBFGS}) for the same network parameters. The convergence of the loss value is much slower requiring almost $10$ times more iteration. Therefore in all the cases we use \emph{Adam} and \emph{BFGS} to train the network and study the minimization problem.

\subsubsection{Case - III~( $2.5\le p \le 5$ and $0 \le \delta \le 0.75$) } 

In the final case we choose right hand side and the boundary conditions so that the exact solution remain same $\mathcal{U}(x,y)=1-x^2-y^2$. We have also consider the same functionals of interest as in Case-II. In addition our neural network structures are also remain same. 

\begin{figure}[htbp]
\centering
\includegraphics[width=3.1in,height=2.3in]{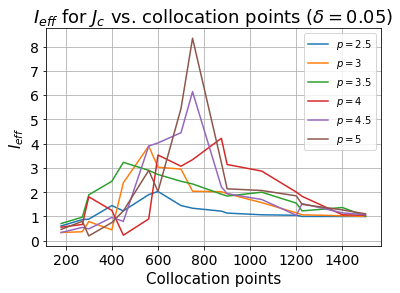}\qquad
\includegraphics[width=3.1in,height=2.3in]{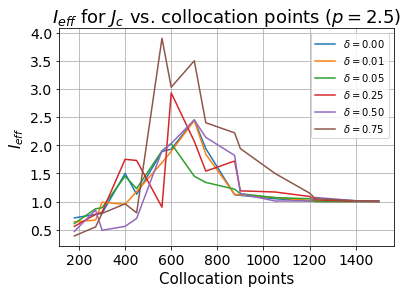}
\caption{$I_{\text{eff}}$ vs. Collocation points}
\label{fig10}
\end{figure}  
As like preceding cases the interesting part is to observe the graph of effectivity index w.r.t increasing collocation points. As it can be seen in Fig.~(\ref{fig10}) how $J_c$ changes its behavior depending on the parameters $(p,\delta)$. With regard to the error we observe that we hardly achieve any advantage at the beginning, $J_c$ even deliver a worse result at a certain step, for relatively higher $p$ and $\delta$.  However at one specific refinement step we start to obtain a better decrease. The poor convergence of $J_c$ might result from the fact that $(p\ge 3, \delta \ge 0.25)$ requires much more refines on domain. Nevertheless in all the cases the convergence rate improves drastically  and we obtain quite good effectivity index once we have $> 1000$ collocation points.


\begin{figure}[htbp]
\centering
\includegraphics[width=3.1in,height=2.3in]{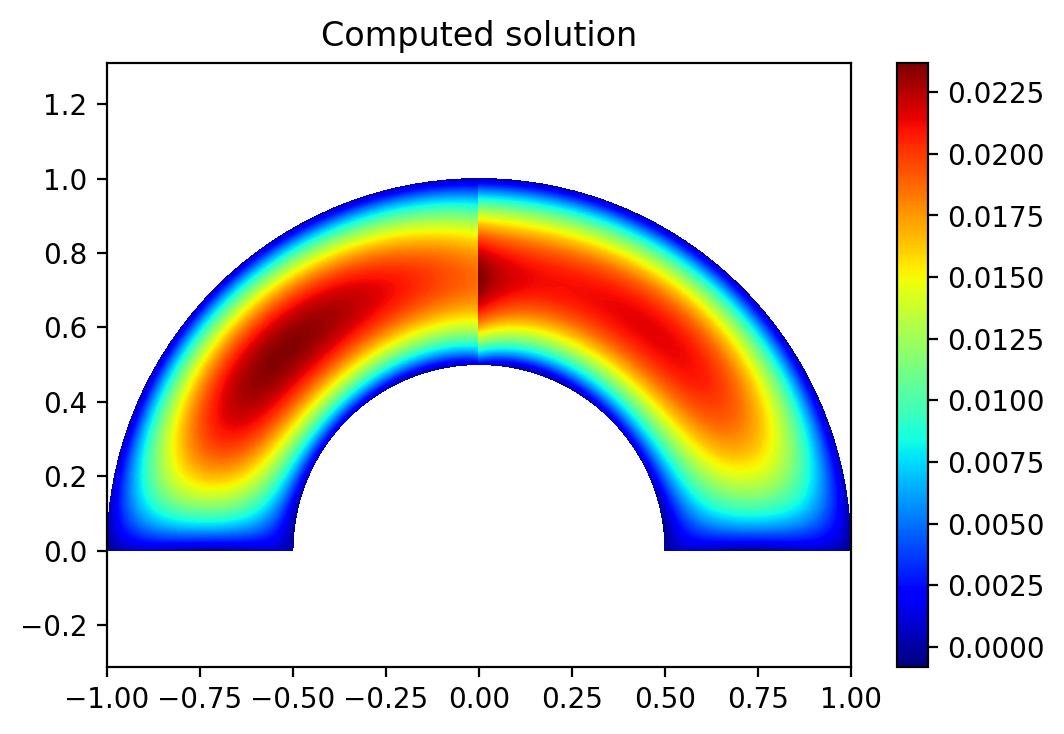}\qquad
\includegraphics[width=3.1in,height=2.3in]{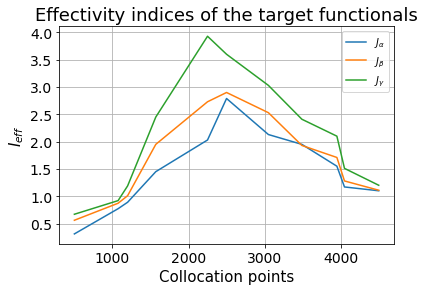}\\
\includegraphics[width=3.1in,height=2.3in]{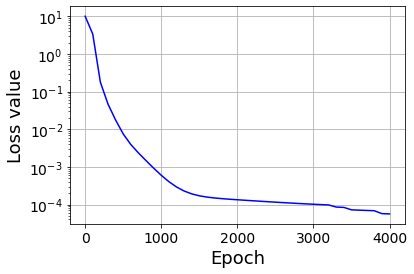}\quad
\includegraphics[width=3.1in,height=2.3in]{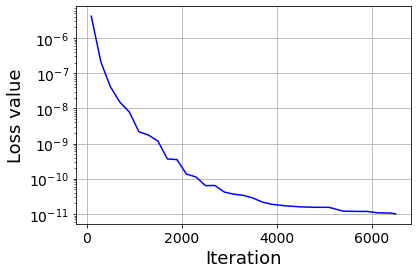}
\caption{Approximated primal solution, $I_{\text{eff}}$ plots and training loss history.}
\label{fig12}
\end{figure}  

Define,
\begin{align*}
  \Omega_{(a,b)}:= \Omega \cap \{(x,y) \in \mathbb{R}^2~:~ ax > by\}  
\end{align*}
\\
We consider the following functionals,

\begin{align*}
    J_1(u) & : = \int_{\Omega_{(0.75,0.5)}} u(x,y) \,dx \,dy \approx 0.003513662 \pm 5 \cdot 10^{-9} \\
    J_2(u) & : = \int_{\Omega_{(-0.75,0.5)}} \Delta u(x,y) \,dx \,dy \approx 0.166887181 \pm 5 \cdot 10^{-9}
\end{align*}

Since the exact solution is unknown to us, we approximated the exact functionals by values that have been obtained by training over $10\, 000$ collocation points and computed the integrals employing Gaussian quadrature over $7\, 500$ points, in order to achieve higher order accuracy. Furthermore, we used three different configurations to formulate goal functionals:
 \begin{itemize}
    \item {Configuration \RNum{1} : evaluating $J_{\alpha}= 10^{-3}\mu_1 \cdot J_1(u) + 0.1 \mu_2 \cdot J_2(u)$ }
    \item{Configuration \RNum{2} : evaluating $J_{\beta}= 0.01\mu_1 \cdot J_1(u) + 0.25 \mu_2 \cdot J_2(u)$ }
     \item{Configuration \RNum{3} : evaluating $J_{\gamma}= 0.09\mu_1 \cdot J_1(u) + 0.5 \mu_2 \cdot J_2(u)$ }
\end{itemize}

Even for a single goal functional the existing literature is rare. Therefore our results might serve a kind of benchmark for comparing algorithms related such problems. The effectivity index for all these configurations are depicted in Fig~(\ref{fig12}). Initially we observe that $\eta$ was dominated by $e$ and later we  get some increases $I_{eff}>1$ for all configurations  even for larger amount of  training data points, which shows that high inconsistency in the decrease of error and because of the loss of regularity in the domain it is not as close to $1$ like earlier examples. However once the domain is refined with more data points, we obtain much better reduction for all configurations. An interesting aspects in this example is that for all the goal functionals we are obtaining almost similar order of reductions in the error. Consequently we notice that our refinement scheme also takes care of less regular problems while accurately approximating the error estimator as far as possible as displayed in the figures.
In the  final example, we consider the steady state Navier Stokes Equations. 


The problem is motivated from \cite{kovasznay1948laminar}.
The domain is given as

\begin{align*}
  \Omega:=\left \{x: \frac{1}{2} <|x| < 1 \right \} \times [-1,1] ~\bigcup ~[-1,1] \times \left \{y: \frac{1}{2} < |y| <1 \right \}
 \end{align*}

\subsection{Example-$3$: Stationary incompressible Navier-Stokes equations} 

\begin{figure}[htbp]
\centering
\includegraphics[width=3in,height=2in]{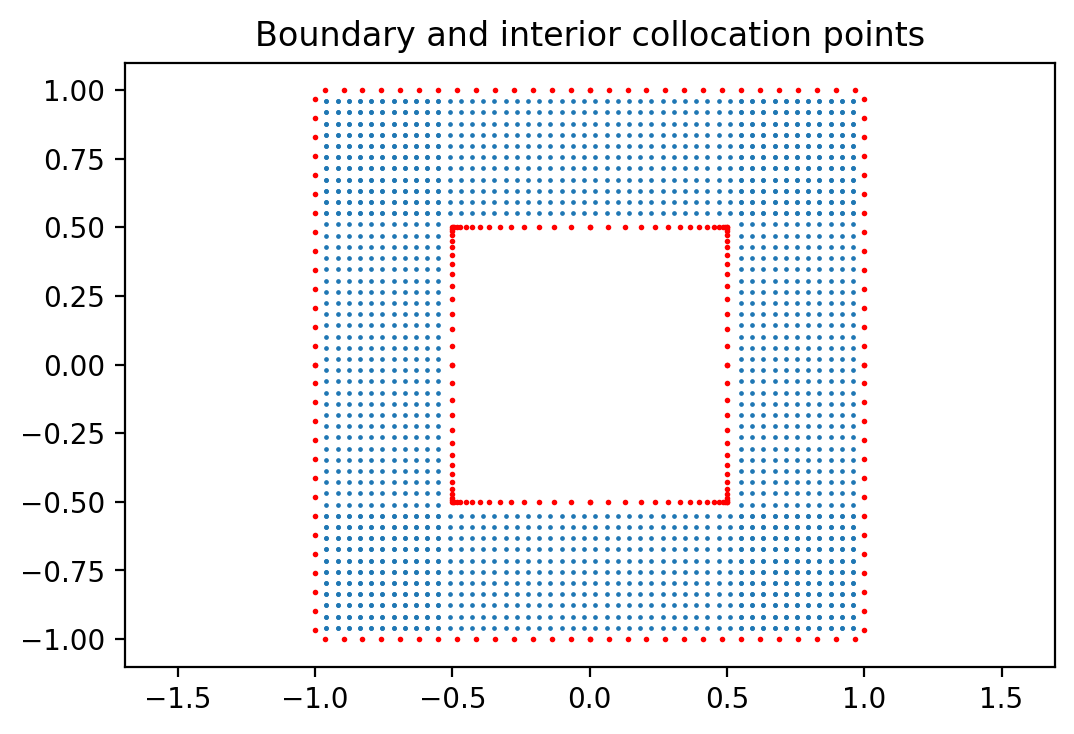}
\caption{Uniformly distributed training points over $\Omega$.}
\label{fig13}
\end{figure}    
In $\Omega$, the incompressible Navier-Stokes equations for a $2$D velocity vector field $\bold{u}(x,y)=[u(x,y)~~v(x,y)]$ and a scalar pressure field $p(x,y)$ are given by
\begin{align*}
  -\nu \Delta \bold{u}+\bold{u} \cdot \nabla \bold{u} + \nabla p  &= \bold{f}~~~~\text{in } \Omega,\\
  \nabla \cdot \bold{u} & = 0 ~~~~\text{in } \Omega,\\
  \bold{u} &= \bold{\psi}~~~~\text{on }\partial \Omega=\Gamma.
\end{align*}

\begin{figure}[htbp]
\centering
\includegraphics[width=2in,height=1.5in]{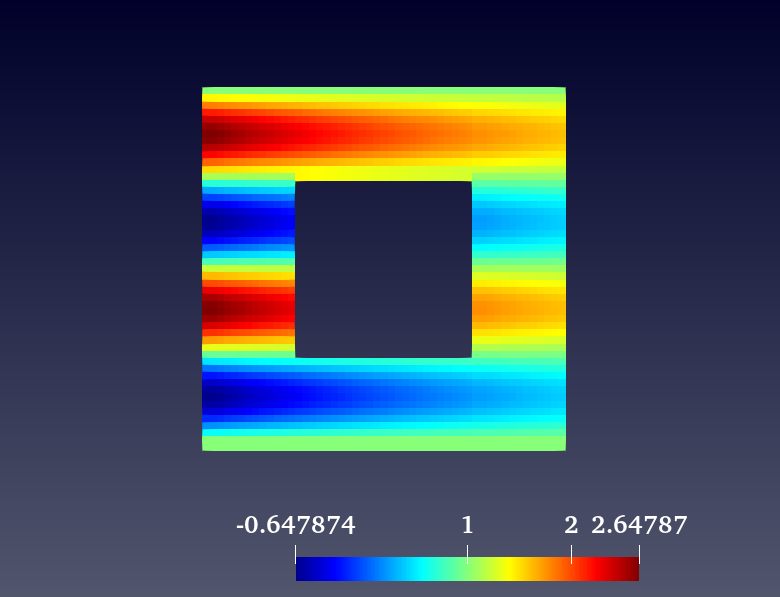}\quad
\includegraphics[width=2in,height=1.5in]{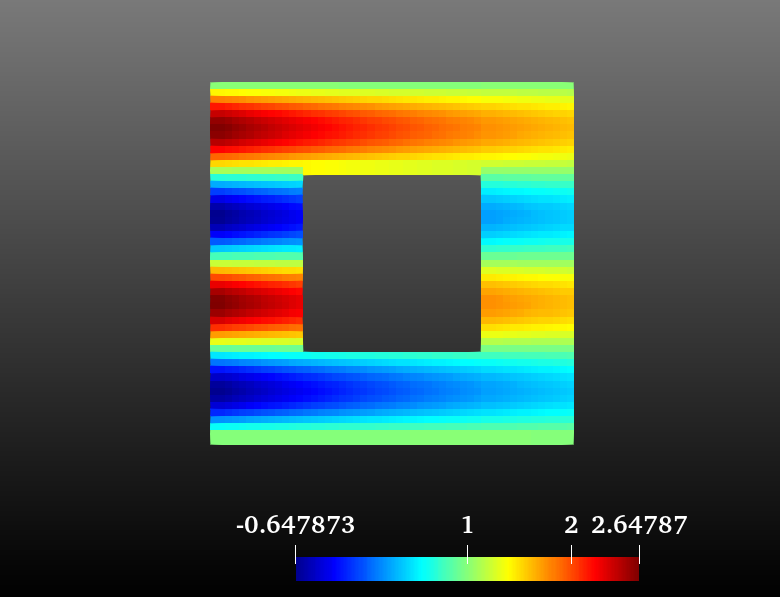}\quad
\includegraphics[width=2in,height=1.5in]{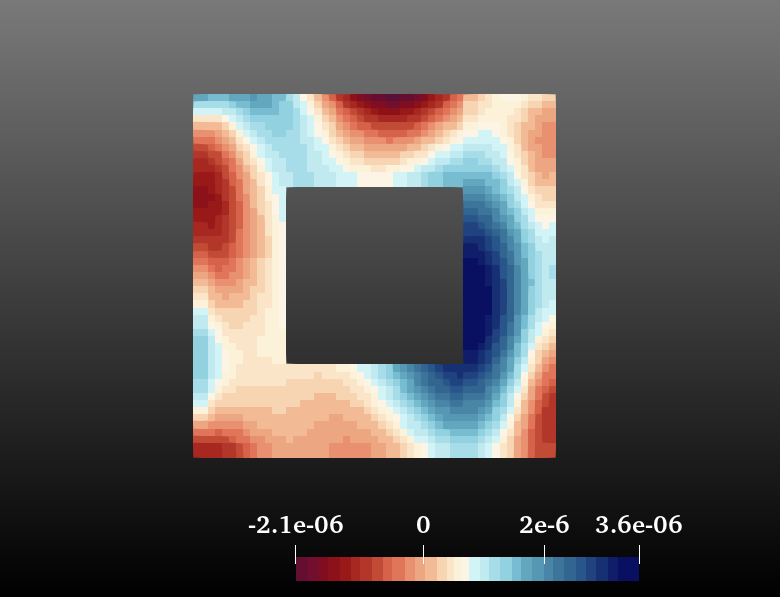}
\caption{Exact solution $(U_{\text{Exact}})$ , Computed solution $(U_{\theta})$ and Point-wise error differences $(U_{\text{Exact}}-U_{\theta})$.}
\label{fig14}
\end{figure}

\begin{figure}[htbp]
\centering
\includegraphics[width=2in,height=1.5in]{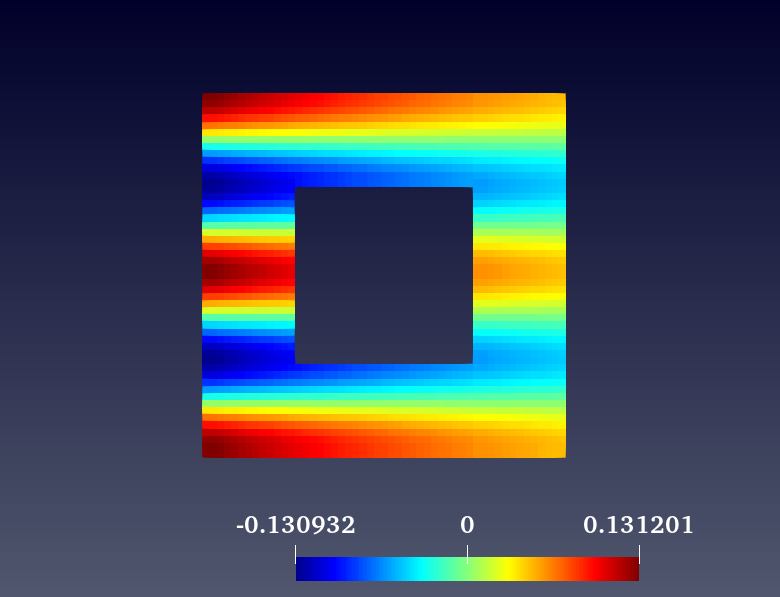}\quad
\includegraphics[width=2in,height=1.5in]{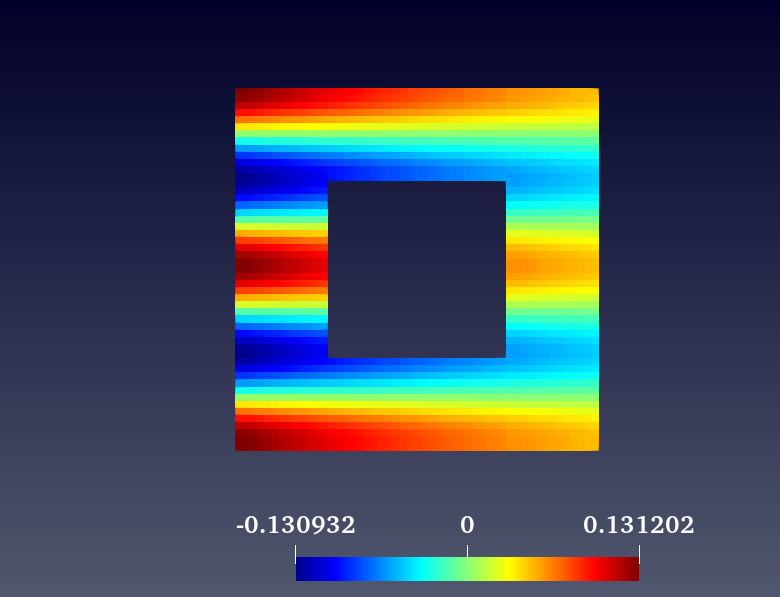}\quad
\includegraphics[width=2in,height=1.5in]{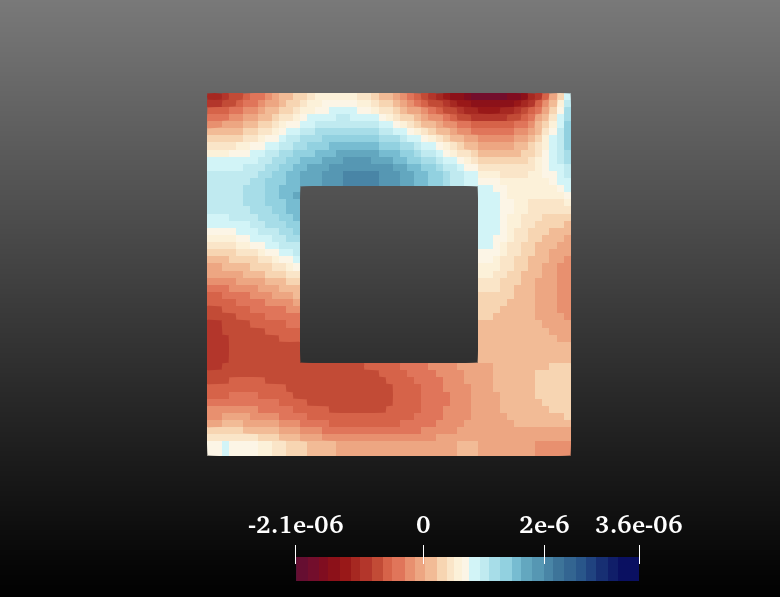}
\caption{Exact solution $(V_{\text{Exact}})$ , Computed solution $(V_{\theta})$ and Point-wise error differences  $(V_{\text{Exact}}-V_{\theta})$.}
\label{fig15}
\end{figure}

\begin{figure}[htbp]
\centering
\includegraphics[width=2in,height=1.5in]{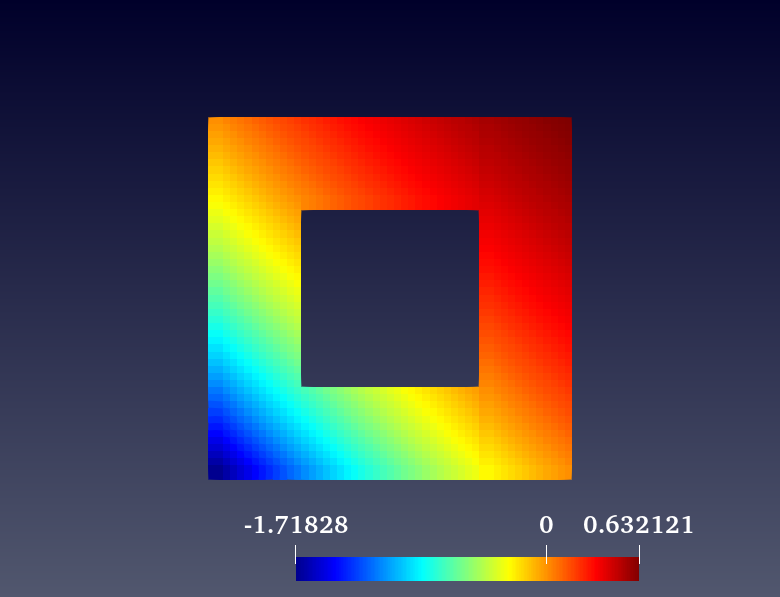}\quad
\includegraphics[width=2in,height=1.5in]{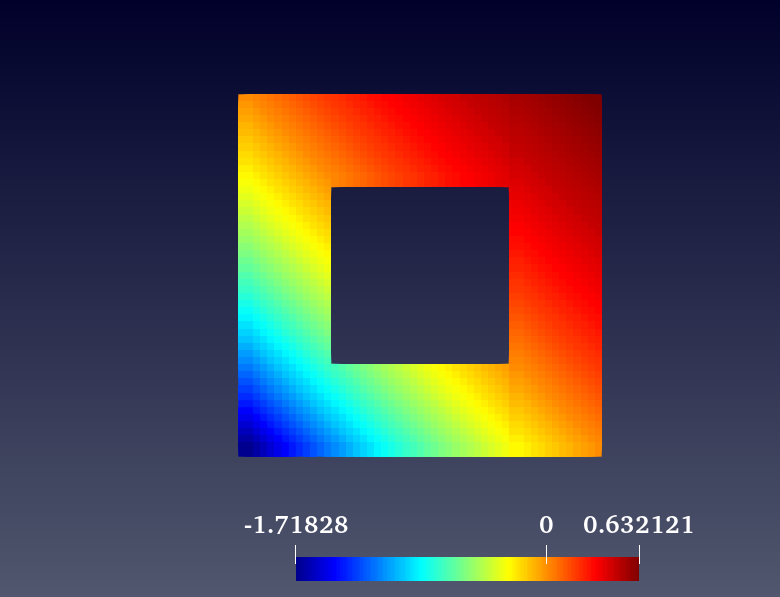}\quad
\includegraphics[width=2in,height=1.5in]{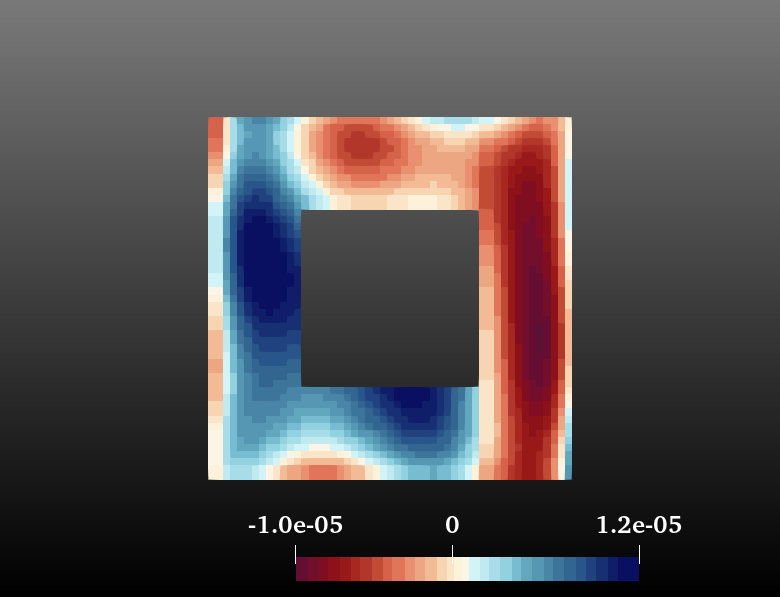}
\caption{Exact solution $(P_{\text{Exact}})$ , Computed solution $(P_{\theta})$ and Point-wise error differences $(P_{\text{Exact}}-P_{\theta})$.}
\label{fig16}
\end{figure}

Here $\nu$ is constant dynamic viscosity of the fluid. We choose $\nu=0.05$. As an exact solution to compare with, we use
\begin{align*}
    u(x,y) & = 1-e^{-\frac{x}{2}} \sin 2 \pi y\\
    v(x,y) & = \frac{1}{4 \pi} e^{-\frac{x}{2}} \cos 2 \pi y\\
    p(x,y) & = 1-e^{-\frac{x+y}{2}}
\end{align*}
and plug it into the original equation to compute $\bold{f}$ and $\bold{\psi}$. The loss function is computed as 
\begin{align*}
    \mathcal{E}:= \frac{1}{N_{int}} \sum_{i=1}^{N_{int}} \sum_{j=1}^2 (\nu \Delta U_{\theta}-U_{\theta} \cdot \nabla U_{\theta} - & \nabla P_{\theta}+ \bold{f})^2_j(\bold{x}_i) + \frac{1}{N_{int}} \sum_{i=1}^{N_{int}}(\nabla \cdot U_{\theta})^2(\bold{x}_i)\\
    & + \frac{1}{N_{bnd}}\sum_{i=1}^{N_{bnd}}\sum_{j=1}^2 (U_{\theta}-\psi)^2_j(\bold{s}_i)
\end{align*}
We generate a dataset containing $5600~ (N_x=80,N_y=70)$ samples of training data see Fig.~(\ref{fig13}). The network architecture used here consists of $6$ layers $(2-80-80-80-80-3)$, \i.e, we have trained one neural network with $3$ outputs. One can also train individually to approximate $\bold{u}$ and $p$ but note that training different architecture is hard because obtaining optimal hyperparameters is itself a daunting task and also it requires lot of computation time. In addition former is relatively better in capturing the underlying equations. As mentioned earlier, the weights are initialized using a Xavier initialization while the biases are generated using a normal distribution with mean $0$ and standard deviation $1$. They are trained using an SGD optimizer with the fine tuned learning rate and momentum followed by BFGS. The first three hidden layers are connected with \enquote{swish} activation and the final layer is with \enquote{tanh}  activation. The resulting prediction error is validated against the test data, Fig.~(\ref{fig14}-\ref{fig16}) provides a comparison between exact solution and predicted outcome. It can be observed that predicted outputs and exact solutions are very close on the testing sets, which clearly shows that using only a handful amount of training  data our model is capable to accurately capture the intricate non linear behavior of N-S equation. The tolerance value  $Tol=10^{-12}$ is adopted as a convergence criterion. In particular Fig.~(\ref{fig17}) summarizes the assessment of our architecture on this dataset.
\subsubsection*{Functionals of Interest} 
\begin{figure}[htbp]
\centering
\includegraphics[width=2in,height=1.5in]{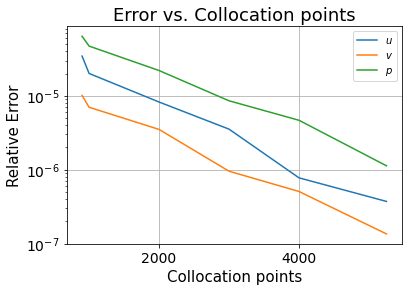}\quad
\includegraphics[width=2in,height=1.5in]{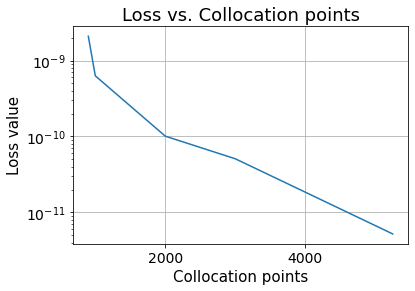}\quad
\includegraphics[width=2in,height=1.5in]{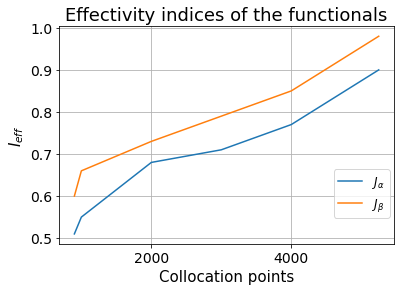}
\caption{Comparison of relative errors, loss values and effectivity indices w.r.t training points.}
\label{fig17}
\end{figure}

We consider the following quantities of interest simultaneously using the framework of multiple goal functionals
\begin{align*}
    J_1(\bold{u}):= c_{Re} \int_{\Gamma} \bold{n} \cdot \sigma \begin{bmatrix}
1 \\
0
\end{bmatrix} \,ds \quad J_2(\bold{u}):=  \pi \int_{\Omega} \left |1-u(x,y) \right | \,d(x,y) \quad J_4(\bold{u}):=  c_{Re} \int_{\Gamma} \bold{n}& \cdot \sigma \begin{bmatrix}
0 \\
1
\end{bmatrix}  \,ds\\
     J_3(\bold{u}):= u(0.75,0.25)+u(-0.75,-0.25) \quad J_5(\bold{u}):=  \pi^2 \int_{\Omega} \left |v(x,y) \right | \,d(x,y) \quad J_6(\bold{u})&:= v(-0.75,0.5)
     \end{align*}
where, $\sigma = -pI_2 + \nu \frac{\nabla \bold{u}+ \nabla \bold{u}^t}{2}$  is the Cauchy stress tensor for the Newtonian fluid, the constant $c_{Re}=\frac{1}{\nu Re}$ is depending upon Reynolds number $Re=100$ and $\bold{n}$ is the outer unit normal to $\Gamma$. In particular the scalar quantities $J_1$ and $J_4$ are known as \emph{drag} and \emph{lift}  coefficients of some body immersed in the flow along the downwind and crosswind directions. We consider the goal functionals :

\begin{align*}
    J_{\alpha}(\bold{u}) & = 0.5 \mu_1J_1(\bold{u})+0.75\mu_2J_2(\bold{u})+\mu_3 J_3(\bold{u})\\
    \\
    J_{\beta}(\bold{u}) & =  0.1 \mu_4 J_4(\bold{u})+0.45 \mu_5 J_5(\bold{u})+0.6\mu_6 J_6(\bold{u})
\end{align*}
The adjoint problem in strong form is given by
\begin{align*}
  -\nu \Delta \bold{z} -\bold{u} \cdot \nabla \bold{z}-(\nabla \cdot \bold{u}) \bold{z}+(\nabla \bold{u})\bold{z} +\nabla \pi= J(\bold{u})  
\end{align*}
For  more detailed descriptions on this topic readers are referred to ~\cite{goll2017dopelib,becker2000optimal,bangerth2003adaptive}. 
Other results for multigoal-oriented error estimation using the 
incompressible Navier-Stokes equations can be found in \cite{EndtLaWi20}. Next, we discuss our findings.
A quantitative behavior of the error plots are displayed in Fig.~(\ref{fig17}). Here again we observe nearly perfect agreement of the estimator with true error yielding the effectivity indices closer to $1$. Evidently our results indicate that the errors converge w.r.t a larger amount of training points.

\section{Conclusions}
\label{sec_conclusions}

In this work, we explored DNN to develop a posteriori error estimation using the dual weighted residual approach for multiple goal functionals. 
Our developments include 
both nonlinear PDEs and nonlinear goal functionals. In the error 
estimator, both the primal and adjoint solutions are approximated 
with the help of neural networks. This procedure is based on the deep collocation method. Moreover, we utilized NURBS for 
constructing the computational domain and Gaussian quadrature rules for an efficient approximation of numerical integration. We remain faithful to the philosophy of using gradient based optimization methods which also allows us to use standard libraries from the machine learning community. One of the principle advantage of our proposed  approach is that a wealth of tools and concept developed by an active community used, that leads ease to implementation, trying different mathematical models and implementing them in straightforward and  transparent manner. We have designed a well-tested framework for evaluating multiple goal functionals for the $2$D $p$-Laplacian and stationary incompressible Navier-Stokes problem, however a more detailed study on the convergence of neural networks and investigating robust minimization procedures still remain open as possible research topics. 
In addition, the extension to evaluate the adjoint error part $\rho^{\ast}$ 
might be of future interest for highly nonlinear problems.
 
\section*{Acknowledgements} 
  This work is supported by the ERC Starting Grant (802205) and the Deutsche Forschungsgemeinschaft (DFG) 
under Germany’s Excellence Strategy within the cluster of Excellence PhoenixD (EXC 2122, Project ID 390833453). The authors acknowledge the support from Dr.Cosmin Anitescu (Weimar) and Dr.Hongwei Guo (Hannover) while carrying out this work.

\bibliographystyle{abbrv}  
\bibliography{references}

\begin{thebibliography}{1}


\bibitem{hughes2012finite}
T.~J. Hughes, The finite element method: linear static and dynamic finite
  element analysis, Courier Corporation, 2012.

\bibitem{rabczuk2019extended}
T.~Rabczuk, J.-H. Song, X.~Zhuang, C.~Anitescu, Extended finite element and
  meshfree methods, Academic Press, 2019.

\bibitem{chakraborty2019finite}
A.~Chakraborty, B.~R. Kumar, Finite element method for drifted space fractional
  tempered diffusion equation, Journal of Applied Mathematics and Computing
  61~(1-2) (2019) 117--135.

\bibitem{jia2019adaptive}
Y.~Jia, C.~Anitescu, Y.~J. Zhang, T.~Rabczuk, An adaptive isogeometric analysis
  collocation method with a recovery-based error estimator, Computer Methods in
  Applied Mechanics and Engineering 345 (2019) 52--74.


\bibitem{abadi2016tensorflow}
Abadi, Martín, et al. Tensorflow: A system for large-scale machine learning.
12th {USENIX} symposium on operating systems design and implementation ({OSDI} 16). 2016.

\bibitem{paszke2017automatic}
A.~Paszke, S.~Gross, S.~Chintala, G.~Chanan, E.~Yang, Z.~DeVito, Z.~Lin,
  A.~Desmaison, L.~Antiga, A.~Lerer, Automatic differentiation in pytorch
  (2017).

\bibitem{samaniego2020energy}
E.~Samaniego, C.~Anitescu, S.~Goswami, V.~M. Nguyen-Thanh, H.~Guo, K.~Hamdia,
  X.~Zhuang, T.~Rabczuk, An energy approach to the solution of partial
  differential equations in computational mechanics via machine learning:
  Concepts, implementation and applications, Computer Methods in Applied
  Mechanics and Engineering 362 (2020) 112790.

\bibitem{sirignano2018dgm}
J.~Sirignano, K.~Spiliopoulos, Dgm: A deep learning algorithm for solving
  partial differential equations, Journal of computational physics 375 (2018)
  1339--1364.

\bibitem{raissi2017physics}
M.~Raissi, P.~Perdikaris, G.~E. Karniadakis, Physics informed deep learning
  (part i): Data-driven solutions of nonlinear partial differential equations,
  arXiv preprint arXiv:1711.10561 (2017).

\bibitem{mishra2018machine}
S.~Mishra, A machine learning framework for data driven acceleration of
  computations of differential equations, arXiv preprint arXiv:1807.09519
  (2018).

\bibitem{brunton2016discovering}
S.~L. Brunton, J.~L. Proctor, J.~N. Kutz, Discovering governing equations from
  data by sparse identification of nonlinear dynamical systems, Proceedings of
  the national academy of sciences 113~(15) (2016) 3932--3937.

\bibitem{rudy2017data}
S.~H. Rudy, S.~L. Brunton, J.~L. Proctor, J.~N. Kutz, Data-driven discovery of
  partial differential equations, Science Advances 3~(4) (2017) e1602614.

\bibitem{schaeffer2017learning}
H.~Schaeffer, Learning partial differential equations via data discovery and
  sparse optimization, Proceedings of the Royal Society A: Mathematical,
  Physical and Engineering Sciences 473~(2197) (2017) 20160446.

\bibitem{brevis2020}
Brevis, Ignacio, Ignacio Muga, and Kristoffer G. van der Zee: 
A machine learning minimal residual (ML MRes) framework for goal oriented finite element discretizations.
Computers and  Mathematics with Applications 95 (2021). 

\bibitem{xu2019dl}
H.~Xu, H.~Chang, D.~Zhang, Dl-pde: Deep-learning based data-driven discovery of
  partial differential equations from discrete and noisy data, arXiv preprint
  arXiv:1908.04463 (2019).

\bibitem{long2018pde}
Z.~Long, Y.~Lu, X.~Ma, B.~Dong, Pde-net: Learning pdes from data, in:
  International Conference on Machine Learning, PMLR, 2018, pp. 3208--3216.

\bibitem{berg2019data}
J.~Berg, K.~Nystr{\"o}m, Data-driven discovery of pdes in complex datasets,
  Journal of Computational Physics 384 (2019) 239--252.

\bibitem{beck2019machine}
C.~Beck, E.~Weinan, A.~Jentzen, Machine learning approximation algorithms for
  high-dimensional fully nonlinear partial differential equations and
  second-order backward stochastic differential equations, Journal of Nonlinear
  Science 29~(4) (2019) 1563--1619.

\bibitem{chaudhari2018deep}
P.~Chaudhari, A.~Oberman, S.~Osher, S.~Soatto, G.~Carlier, Deep relaxation:
  partial differential equations for optimizing deep neural networks, Research
  in the Mathematical Sciences 5~(3) (2018) 30.


\bibitem{BeRa96}
R.~Becker, R.~Rannacher, A feed-back approach to error control in finite
  element methods: basic analysis and examples, East-West J. Numer. Math. 4
  (1996) 237--264.

\bibitem{becker2001optimal}
R.~Becker, R.~Rannacher, An optimal control approach to a posteriori error
  estimation in finite element methods, Acta numerica 10~(1) (2001) 1--102.

\bibitem{AinsworthOden:2000}
M.~Ainsworth, J.~T. Oden, A Posteriori Error Estimation in Finite Element
  Analysis, Pure and Applied Mathematics (New York), Wiley-Interscience [John
  Wiley \& Sons], New York, 2000.

\bibitem{bangerth2003adaptive}
W.~Bangerth, R.~Rannacher, Adaptive finite element methods for differential
  equations, Birkh{\"a}user, 2003.


\bibitem{RiWi15_dwr}
T.~Richter, T.~Wick, Variational localizations of the dual weighted residual
  estimator, Journal of Computational and Applied Mathematics 279~(0) (2015)
  192 -- 208.

\bibitem{Od18}
J.~T. Oden, Adaptive multiscale predictive modelling, Acta Numerica 27 (2018)
  353–450.
\newblock \href {https://doi.org/10.1017/S096249291800003X}
  {\path{doi:10.1017/S096249291800003X}}.

\bibitem{EndtLaWi20}
B.~Endtmayer, U.~Langer, T.~Wick,
  \href{https://doi.org/10.1137/18M1227275}{Two-{S}ide a {P}osteriori {E}rror
  {E}stimates for the {D}ual-{W}eighted {R}esidual {M}ethod}, SIAM J. Sci.
  Comput. 42~(1) (2020) A371--A394.
 \href{https://doi.org/10.1137/18M1227275}
  {\path{doi:10.1137/18M1227275}}.
\url{https://doi.org/10.1137/18M1227275}

\bibitem{Wi20_book}
T.~Wick, \href{https://www.degruyter.com/view/title/523232}{Multiphysics
  Phase-Field Fracture: Modeling, Adaptive Discretizations, and Solvers}, De
  Gruyter, Berlin, Boston, 2020.
\href{https://doi.org/https://doi.org/10.1515/9783110497397}
  {\path{doi:https://doi.org/10.1515/9783110497397}}.
\url{https://www.degruyter.com/view/title/523232}

\bibitem{hartmann2003goal}
R.~Hartmann, P.~Houston, Goal-oriented a posteriori error estimation for
  multiple target functionals, in: Hyperbolic problems: theory, numerics,
  applications, Springer, 2003, pp. 579--588.

\bibitem{hartmann2008multitarget}
R.~Hartmann, Multitarget error estimation and adaptivity in aerodynamic flow
  simulations, SIAM Journal on Scientific Computing 31~(1) (2008) 708--731.


\bibitem{endtmayer2017partition}
B.~Endtmayer, T.~Wick, A partition-of-unity dual-weighted residual approach for
  multi-objective goal functional error estimation applied to elliptic
  problems, Computational Methods in Applied Mathematics 17~(4) (2017)
  575--599.

\bibitem{endtmayer2019multigoal}
B.~Endtmayer, U.~Langer, T.~Wick, Multigoal-oriented error estimates for
  non-linear problems, Journal of Numerical Mathematics 27~(4) (2019) 215--236.

\bibitem{Endt21}
B.~Endtmayer, Multi-goal oriented a posteriori error estimates for nonlinear
  partial differential equations, Ph.D. thesis, Johannes Kepler University Linz
  (2021).


\bibitem{braackern2003}
M.~Braack, A.~Ern, \href{https://doi.org/10.1137/S1540345902410482}{A
  posteriori control of modeling errors and discretization errors}, Multiscale
  Modeling \& Simulation 1~(2) (2003) 221--238.
\href{http://arxiv.org/abs/https://doi.org/10.1137/S1540345902410482}
  {\path{arXiv:https://doi.org/10.1137/S1540345902410482}}, 
 \href{https://doi.org/10.1137/S1540345902410482}
  {\path{doi:10.1137/S1540345902410482}}.
\url{https://doi.org/10.1137/S1540345902410482}



\bibitem{roth2021neural}
J.~Roth, M.~Schr{\"o}der, T.~Wick, Neural network guided adjoint computations
  in dual weighted residual error estimation (arXiv:2102.12450, 2021).

\bibitem{scheithauerjorge}
J.~Nocedal, S.~J. Wright, Numerical Optimization, Springer, 2006.


\bibitem{saha2019evolution}
S.~Saha, N.~Nagaraj, A.~Mathur, R.~Yedida, Evolution of novel activation
  functions in neural network training with applications to classification of
  exoplanets, arXiv preprint arXiv:1906.01975 (2019).

\bibitem{glorot2010understanding}
X.~Glorot, Y.~Bengio, Understanding the difficulty of training deep feedforward
  neural networks, in: Proceedings of the thirteenth international conference
  on artificial intelligence and statistics, 2010, pp. 249--256.

\bibitem{braack_ern_2003}
M.~Braack, A.~Ern, \href{https://doi.org/10.1137/S1540345902410482}{A
  posteriori control of modeling errors and discretization errors}, Multiscale
  Modeling \& Simulation 1~(2) (2003) 221--238.
 \href{http://arxiv.org/abs/https://doi.org/10.1137/S1540345902410482}
  {\path{arXiv:https://doi.org/10.1137/S1540345902410482}}, \href
  {https://doi.org/10.1137/S1540345902410482}
  {\path{doi:10.1137/S1540345902410482}}.
\url{https://doi.org/10.1137/S1540345902410482}


\bibitem{Cia87}
P.~G. Ciarlet, The finite element method for elliptic problems, 2nd Edition,
  North-Holland, Amsterdam [u.a.], 1987.

\bibitem{cottrell2009isogeometric}
J.~A. Cottrell, T.~J. Hughes, Y.~Bazilevs, Isogeometric analysis: toward
  integration of CAD and FEA, John Wiley \& Sons, 2009.

\bibitem{nguyen2015isogeometric}
V.~P. Nguyen, C.~Anitescu, S.~P. Bordas, T.~Rabczuk, Isogeometric analysis: an
  overview and computer implementation aspects, Mathematics and Computers in
  Simulation 117 (2015) 89--116.

\bibitem{hughes2005isogeometric}
T.~J. Hughes, J.~A. Cottrell, Y.~Bazilevs, Isogeometric analysis: Cad, finite
  elements, nurbs, exact geometry and mesh refinement, Computer methods in
  applied mechanics and engineering 194~(39-41) (2005) 4135--4195.

\bibitem{piegl1996nurbs}
L.~Piegl, W.~Tiller, The NURBS book, Springer Science \& Business Media, 1996.

\bibitem{dimas19993d}
E.~Dimas, D.~Briassoulis, 3d geometric modelling based on nurbs: a review,
  Advances in Engineering Software 30~(9-11) (1999) 741--751.

\bibitem{bingol2019nurbs}
O.~R. Bingol, A.~Krishnamurthy, Nurbs-python: An open-source object-oriented
  nurbs modeling framework in python, SoftwareX 9 (2019) 85--94.

\bibitem{kovasznay1948laminar}
L.~Kovasznay, Laminar flow behind a two-dimensional grid, in: Mathematical
  Proceedings of the Cambridge Philosophical Society, Vol.~44, Cambridge
  University Press, 1948, pp. 58--62.

\bibitem{TP}
\href{https://www.tensorflow.org/probability/examples/A_Tour_of_TensorFlow_Probability}{A
  tour of tensorflow probability}.
\url{https://www.tensorflow.org/probability/examples/A_Tour_of_TensorFlow_Probability}


\bibitem{goll2017dopelib}
C.~Goll, T.~Wick, W.~Wollner, Dopelib: Differential equations and optimization
  environment; a goal oriented software library for solving pdes and
  optimization problems with pdes, Archive of Numerical Software 5~(2) (2017).

\bibitem{becker2000optimal}
R.~Becker, An optimal-control approach to a-posteriori error estimation for
  finite element discretizations of the navier-stokes equations (2000).

\bibitem{Keras}
\href{https://www.keras.io/api/models/model/}{Keras model documentation}.
\url{https://www.keras.io/api/models/model/}

\bibitem{KerPruChaLaf2017}
K.~Kergrene, S.~Prudhomme, L.~Chamoin, M.~Laforest,
  \href{http://www.sciencedirect.com/science/article/pii/S0045782517306485}{A
  new goal-oriented formulation of the finite element method}, Computer Methods
  in Applied Mechanics and Engineering 327 (2017) 256 -- 276, advances in
  Computational Mechanics and Scientific Computation—the Cutting Edge.
\href{https://doi.org/https://doi.org/10.1016/j.cma.2017.09.018}
  {\path{doi:https://doi.org/10.1016/j.cma.2017.09.018}}.
\url{http://www.sciencedirect.com/science/article/pii/S0045782517306485}

\bibitem{BruZhuZwie16}
E.~van Brummelen, S.~Zhuk, G.~van Zwieten,
  \href{http://www.sciencedirect.com/science/article/pii/S0045782516301815}{Worst-case
  multi-objective error estimation and adaptivity}, Computer Methods in Applied
  Mechanics and Engineering 313 (2017) 723 -- 743.
\href{https://doi.org/https://doi.org/10.1016/j.cma.2016.10.007}
  {\path{doi:https://doi.org/10.1016/j.cma.2016.10.007}}.
\url{http://www.sciencedirect.com/science/article/pii/S0045782516301815}


\end{thebibliography}

\end{document}